%% file: iclr2016_conference.tex
\documentclass{article} 
\usepackage{iclr2016_conference,times}
\usepackage{hyperref}
\usepackage{url}
\usepackage{graphicx}
\usepackage{amsmath}
\usepackage{amsfonts}
\usepackage{amssymb}
\usepackage{amsthm}
\usepackage{epstopdf}
\usepackage{wrapfig}
\usepackage[hang,small]{caption}
\usepackage{bm}
\usepackage{subcaption}
\usepackage{rotating}

\usepackage{todonotes}

\title{A Test of Relative Similarity for Model\\ Selection in Generative Models}

\author{Wacha Bounliphone,$^{12}$\thanks{These authors contributed equally to this work} ~ Eugene Belilovsky,$^{12}$\footnotemark[1] ~ \&   Matthew B.\ Blaschko$^2$\\
  $^1$CentraleSup\'{e}lec \& Inria Saclay, Universit\'{e} Paris-Saclay,
  92295 Ch\^{a}tenay-Malabry, France\\
  $^2$ESAT-PSI, KU Leuven,  Kasteelpark Arenberg 10, 3001 Leuven, Belgium\\
  \texttt{\{wacha.bounliphone,eugene.belilovsky\}@inria.fr}\\
  \texttt{matthew.blaschko@esat.kuleuven.be}\\
  \And
  Ioannis Antonoglou\\
  Google Deepmind\\
  5 New Street Square\\
  London EC4A 3TW, UK\\
  \texttt{ioannisa@google.com}
  \And
Arthur Gretton \\
Gatsby Computational Neuroscience Unit\\
University College London \\
25 Howland Street\\
London W1T 4JG, UK\\
\texttt{arthur.gretton@gmail.com}
}

\newcommand{\MMDpop}[2]{\operatorname{MMD}(\mathcal{F},P_{#1},P_{#2})}

\newcommand{\squaredMMDpop}[2]{\operatorname{MMD}^2(\mathcal{F},P_{#1},P_{#2})}
\newcommand{\squaredMMDu}[2]{\operatorname{MMD}_u^2(\mathcal{F},{#1},{#2})}

\newcommand\E{\operatorname{E}}
\newcommand\var{\operatorname{Var}}
\newcommand{\Eone}{\operatorname{E}_{v_1}}

\newcommand{\EXoneYone}{\operatorname{E}_{x_1, y_1}}
\newcommand{\EXtwoYtwo}{\operatorname{E}_{x_2,y_2}}
\newcommand{\EXone}{\operatorname{E}_{x_1}}
\newcommand{\EYone}{\operatorname{E}_{y_1}}
\newcommand{\EZone}{\operatorname{E}_{z_1}}
\newcommand{\EXoneYoneZone}{\operatorname{E}_{x_1,y_1,z_1}}
\newcommand{\EXtwoYtwoZtwo}{\operatorname{E}_{x_2, y_2,z_2}}

\newtheorem{Pb}{Problem}
\newtheorem{Thm}{Theorem}
\newtheorem{Def}{Definition}

\iclrfinalcopy 

\begin{document}

\maketitle
\input{abstract.tex}
\input{introduction.tex}

\input{sec_definition.tex}
\input{sec_statistictest.tex}

\input{sec_experiments.tex}

\input{sec_model_selection.tex}

\input{AutoEncExpDiscussion.tex}
\input{conclusion.tex}

\subsubsection*{Acknowledgments}

We thank Joel Veness for helpful comments. This work is partially funded by Internal Funds KU Leuven, ERC Grant 259112, FP7-MC-CIG 334380, the Royal Academy of Engineering through the Newton Alumni Scheme, and DIGITEO 2013-0788D-SOPRANO. WB is supported by a CentraleSup\'{e}lec fellowship.

\bibliography{bibliography}
\bibliographystyle{abbrvnat}

\appendix

\input{detailedCovarianceDerivations.tex}
\input{calibrationOftheTest.tex}

\end{document}

%% file: abstract.tex
\begin{abstract}
Probabilistic generative models provide a powerful framework for representing data that avoids the expense of manual annotation typically needed by discriminative approaches.  Model selection in this generative setting can be challenging, however, particularly when likelihoods are not easily accessible.  To address this issue, we introduce a statistical test of relative similarity, which is used to determine which of two models generates samples that are significantly closer to a real-world reference dataset of interest.  We use as our test statistic the difference in maximum mean discrepancies (MMDs) between the reference dataset and each model dataset, and derive a powerful, low-variance test based on the joint asymptotic distribution of the MMDs between each reference-model pair.
In experiments on deep generative models,
including the variational auto-encoder and generative moment matching network, the tests provide a meaningful ranking of model performance as a function of parameter and training settings.
\end{abstract}

%% file: introduction.tex
\section{Introduction} 

Generative models based on deep learning techniques aim to provide sophisticated and accurate models of data, without expensive manual annotation~\citep{bengio2009learning,kingma2014semi}.  This is especially of interest as deep networks tend to require comparatively large training samples to achieve a good result~\citep{krizhevsky2012imagenet}.  Model selection within this class of techniques can be a challenge, however.
First, likelihoods can be difficult to compute for some families of recently proposed models based on deep learning~\citep{goodfellow2014generative,li2015generative}. The current best method to evaluate such models is based on Parzen-window estimates of the log likelihood \cite[Section 5]{goodfellow2014generative}. Second, if we are  given two models with similar likelihoods, we typically do not have a computationally inexpensive hypothesis test to determine whether one likelihood is significantly higher than the other.  Permutation testing or other generic strategies are often computationally prohibitive, bearing in mind the relatively high computational requirements of deep networks~\citep{krizhevsky2012imagenet}.

In this work, we provide an alternative strategy for model selection,
based on a novel, non-parametric hypothesis test of relative similarity.
We treat the two  trained networks being compared as  generative models \citep{goodfellow2014generative,hinton2006fast,salakhutdinov2009deep}, and test whether the first candidate model generates samples significantly closer to a 
reference  validation set.
The null hypothesis is that the ordering is reversed, and the second candidate model is closer to the reference (further, both samples are assumed to remain distinct from the reference, as will be the case for any sufficiently complex modeling problem).


Our model selection criterion is based on the maximum mean discrepancy (MMD) \citep{gretton2006kernel,gretton2012kernel}, which represents the distance between embeddings of empirical distributions in a reproducing kernel Hilbert space (RKHS). The maximum mean discrepancy
is a metric on the space of probability distirbutions when a characteristic kernel is used \citep{FukGreSunSch08,SriGreFukLanetal10}, meaning that the distribution embeddings are unique for each probability measure.
Recently, the MMD has been used in training generative models adversarially, \citep{li2015generative,dziugaite2015training}, where the MMD measures the distance of the generated samples to some reference target set; it has been used for statistical model criticism \citep{lloydstatistical}; and to minimize the effect of nuisance variables on learned representations \citep{louizos2015variational}.

Rather than {\em train} a single model using the MMD distance to a reference distribution, our goal in this work is to  {\em evaluate the relative performance}  of two models, by testing whether one generates samples significantly closer to the reference distribution than the other. This extends the applicability of the MMD to  problems of model selection and evaluation. Key to this result is a novel expression for the joint asymptotic distribution of two correlated MMDs (between samples generated from each model, and samples from the reference distribution).  \cite{li2015m} have derived the joint distribution of a specific MMD estimator under the assumption that the distributions are equal.  By contrast, we derive the case in which the distributions are unequal, as is expected due to irreducible model error.

We provide a detailed introduction to the MMD and its associated notation in Section~\ref{sec:background}.
We derive the joint asymptotic distribution of the MMDs in Section~\ref{sec:theory}: this uses similar ideas to the relative dependence test in \cite{bounliphone2015}, with the additional complexity due to there being three independent samples to deal with, rather than a single joint sample.
We formulate a hypothesis test of {\em relative similarity}, to determine whether the difference in MMDs is statistically significant. Our first test benchmark is on a synthetic data for which the ground truth is known (Section~\ref{sec:ExperimentsBasic}), where we verify that the test performs correctly under the null and the alternative.




Finally, in Section~\ref{sec:ModelSelectionDeepUnsupNN}, 
we demonstrate the performance of our test over a broad selection of model comparison problems in the deep learning setting, by evaluating relative similarity of pairs of model outputs to a validation set over a range of training regimes and settings. Our benchmark models include the variational auto-encoder \citep{kingma2013auto} and the generative moment matching network \citep{li2015generative}. We first demonstrate that the test performs as expected in scenarios where the same model is trained with different training set sizes, and the relative ordering of model performance is known. We then fix the training set size and change various architectural parameters of these networks,  showing which models are significantly preferred with our test. We validate the rankings returned by the test using a separate set of data for which we compute alternate metrics for assessing the models, such as classification accuracy and likelihood.

%% file: sec_definition.tex
\section{Background Material}\label{sec:background}

In comparing samples from distributions, we use the  Maximum Mean Discrepancy ($\operatorname{MMD}$)~\citep{gretton2006kernel,gretton2012kernel}.  We briefly review this statistic and its asymptotic behaviour for a single pair of samples.


\begin{Def}\citep[Definition 2:  Maximum Mean Discrepancy ($\operatorname{MMD}$)]{gretton2012kernel}
 Let $\mathcal{F}$ be an RKHS, with the continuous feature mapping $\varphi(x) \in \mathcal{F}$ from each $x \in \mathcal{X}$, such that the inner product between the features is given by the kernel function $k(x,x') := \langle \phi(x), \phi(x')\rangle$.  Then the squared population $\operatorname{MMD}$ is
\begin{equation}
\squaredMMDpop{x}{y} = \E_{x,x'} \left[ k(x,x') \right] - 2 \E_{x,y} \left[ k(x,y) \right] + \E_{y,y'} \left[ k(y,y') \right]  .
\end{equation}\label{eq:squaredMMDpop}
\end{Def}

The following theorem describes an unbiased quadratic-time estimate of the MMD, and  its asymptotic distribution when $P_x$ and $ P_y$ are different.
\begin{Thm}\citep[Lemma 6 $\&$ Corollary 16: Unbiased empirical estimate $\&$ Asymptotic distribution of $\squaredMMDu{X_m}{Y_n}$]{gretton2012kernel}
Define observations $X_m:= \lbrace x_1, ..., x_m \rbrace$ and $Y_n:=\lbrace y_1, ..., y_n \rbrace$ independently and identically distributed (i.i.d.) from $P_x$ and $P_y$, respectively.
  An unbiased empirical estimate of $\squaredMMDpop{x}{y}$ is a sum of two $U$-statistics and a sample average,
\begin{align}\label{eq:MMDkernelsStuff}
\squaredMMDu{X_m}{Y_n} &= \frac{1}{m(m-1)} \sum_{i=1}^m \sum_{j \ne i}^m k(x_i,x_j) + \frac{1}{n(n-1)} \sum_{i=1}^n \sum_{j \ne i}^n k(y_i,y_j) \\
&- \frac{2}{mn} \sum_{i=1}^m \sum_{j=1}^n k(x_i,y_j). \nonumber
\end{align}
Let $\mathcal{V} := (v_1,...,v_m)$ be $m$ i.i.d. random variables, where $v := (x,y) \sim P_x \times P_y$. When $ m = n$, an unbiased empirical estimate of $\squaredMMDpop{x}{y}$ is 
\begin{equation}
\label{def:def_MMD_stat}
\squaredMMDu{X_m}{Y_m} = \frac{1}{m(m-1)} \sum_{i \ne j}^m h(v_i,v_j)
\end{equation}
with $h(v_i,v_j) = k(x_i,x_j) + k(y_i,y_j) -k(x_i,y_j)  -k(x_j,y_i)$.
We assume $\mathbf{E}(h^2) < \infty$.  When $P_x\neq P_y$, $\squaredMMDu{X}{Y}$ converges in distribution to a Gaussian according to 
\begin{equation}
\sqrt{m} \left( \squaredMMDu{X_m}{Y_n} - \squaredMMDpop{x}{y} \right)  \overset{D}{\longrightarrow} \mathcal{N} \left( 0, \sigma^2_{XY} \right)
\label{eq:asymptotic_distribution_MMD}
\end{equation}
where 
\begin{equation}
\sigma^2_{XY} = 4 \left( \E_{v_1} [(E_{v_2}h(v_1,v_2))^2] - [(E_{v_1,v_2}h(v_1,v_2))^2] \right)
\label{eq:variance_two_MMD}
\end{equation}
uniformly at rate $1/\sqrt{m}$.
\end{Thm}

A two-sample test may be constructed using the MMD as a test statistic, however when $P_x=P_y$ the statistic is degenerate, and
the asymptotic distribution is a weighted sum of  $\chi^2$ variables (which can have infinitely many terms, \citep{gretton2012kernel}).

By contrast, our problem setting is to determine with high significance whether a target distribution $P_x$ is closer to one of two candidate distributions $P_y,P_z$, based on two empirical estimates of the MMD and their variances. This requires us to characterize  $\sigma^2_{XY}$ for $\squaredMMDu{X_m}{Y_n}$ as well as the covariance of two dependent estimates, $\squaredMMDu{X_m}{Y_n}$ and $\squaredMMDu{X_m}{Z_r}$ (the dependence arises from the shared sample $X_m$). Fortunately, degeneracy does not arise if we assume $P_y,P_z$ are each distinct from $P_x$.

In the next section, we obtain the joint asymptotic distribution of two dependent MMD statistics. We demonstrate how this joint distribution can be empirically estimated, and use the resulting parametric form to construct a computationally efficient and powerful hypothesis test for relative similarity.



%% file: sec_statistictest.tex
\section{Joint asymptotic distribution of two correlated MMDs and a resulting test statistic}\label{sec:theory}

In this section, we derive our statistical test for relative similarity as measured by MMD.  In order to maximize the statistical efficiency of the test, we will reuse samples from the reference distribution, denoted by $P_x$, to compute the MMD estimates with two candidate distributions $P_y$ and $P_z$.  
We consider two MMD estimates $\squaredMMDu{X_m}{Y_n}$ and $\squaredMMDu{X_m}{Z_r}$, and as the data sample $X_m$ is identical between them, these estimates will be correlated.  We therefore first derive the joint asymptotic distribution of these two metrics and use this to construct a statistical test.

\begin{Thm}\label{thm:joint_asymptotic_MMD}
We assume that $P_x \ne P_y$, $P_x \ne P_z$,  $E(k(x_i,x_j)) < \infty$,  $E(k(y_i,y_j)) < \infty$ and $E(k(x_i,y_j)) < \infty$, then 
\begin{equation}
\sqrt{m} 
\left( \begin{pmatrix}
\squaredMMDu{X_m}{Y_n} \\ 
\squaredMMDu{X_m}{Z_r} 
\end{pmatrix}
-
\begin{pmatrix}
\squaredMMDpop{x}{y} \\ 
\squaredMMDpop{x}{z}
\end{pmatrix}  
\right)
\overset{d}{\longrightarrow} \mathcal{N} 
\left( 
\begin{pmatrix}
0 \\ 
0
\end{pmatrix},
\begin{pmatrix} 
\sigma_{XY}^2 & \sigma_{XYXZ} \\ 
\sigma_{XYXZ} & \sigma_{XZ}^2 
\end{pmatrix} 
\right)
\label{eq:joint_asymptotic_MMD}
\end{equation}
\end{Thm}

We substitute the kernel MMD definition from Equation~\eqref{eq:MMDkernelsStuff}, expand the terms in the expectation, and determine their empirical estimates in order to compute the variances in practice.  The proof and additional details of the following derivations are given in Appendix~\ref{sec:appendixA}.


An empirical estimate of $\sigma_{XYXZ}$ in Equation~\eqref{eq:joint_asymptotic_MMD}, neglecting higher order terms, can be computed in $\mathcal{O}(m^2)$:
\begin{align}\label{eq:approximation_derivation_of_covariance}
\sigma_{XYXZ} &\approx
\frac{1}{m(m-1)^2} e^T \tilde{K}_{xx}\tilde{K}_{xx} e - \left(\frac{1}{m(m-1)} e^T \tilde{K}_{xx} e \right)^2 \\
& \qquad - \left( \frac{1}{m(m-1)r} e^T \tilde{K}_{xx} K_{xz} e - \frac{1}{m^2(m-1)r}  e^T \tilde{K}_{xx} e e^T K_{xz} e \right) \nonumber \\
& \qquad - \left( \frac{1}{m(m-1)n} e^T \tilde{K}_{xx} K_{xy} e - \frac{1}{m^2(m-1)n} e^T \tilde{K}_{xx} e e^T K_{xz} e \right) \nonumber \\
& \qquad + \left( \frac{1}{mnr} e^T K_{yx} K_{xz} e - \frac{1}{m^2nr} e^T K_{xy}e e^T K_{xz} e\right) \nonumber
\end{align}

where $e$ is a vector of $1$s with appropriate size, while $\tilde{K}_{xx},K_{xy}$ and $K_{xz}$ refer to the kernel matrices, with $\tilde{K}$ indicating that the diagonal entries have been set to zero (cf.\ Appendix~\ref{sec:appendixA}). Similarly, Equation~\eqref{eq:variance_two_MMD} is constructed as in Equation~\eqref{eq:approximation_derivation_of_covariance}.

Based on the empirical distribution from Equation~\eqref{eq:joint_asymptotic_MMD}, we now describe a statistical test to solve the following problem:
 
\begin{Pb}[Relative similarity test]\label{pb:theproblem}
Let $P_x, P_y, x$ and $y$ be defined as above, $z$ be an independent random variables with distribution $P_z$. Given observations $X_m:= \lbrace x_1, ..., x_m \rbrace$, $Y_n:=\lbrace y_1, ..., y_n \rbrace$ and $Z_r:=\lbrace z_1, ..., z_r \rbrace$ i.i.d. from $P_x$, $P_y$ and $P_z$ respectively such that $P_x \ne P_y$, $P_x \ne P_z$, we test the hypothesis that $P_x$ is closer to $P_z$ than $P_y$ i.e. we test the null hypothesis $\mathcal{H}_0$: $ \MMDpop{x}{y} \leq \MMDpop{x}{z}$ versus the alternative hypothesis $\mathcal{H}_1$: $\MMDpop{x}{y} > \MMDpop{x}{z}$ at a given significance level $\alpha$
\end{Pb} 
  
The test statistic $\squaredMMDu{X_m}{Y_n}-\squaredMMDu{X_m}{Z_r}$ is used to compute the $p$-value $p$ for the standard normal distribution.  The test statistic is obtained by rotating the joint distribution (cf.\ Eq.~\ref{eq:joint_asymptotic_MMD}) by $\frac{\pi}{4}$ about the origin, and integrating the resulting projection on the first axis, in a manner similar to~\citet{bounliphone2015}.  Denote the asymptotically normal distribution of $\sqrt{m}[\squaredMMDu{X_m}{Y_n};\: \squaredMMDu{X_m}{Z_r}]^T$ as $\mathcal{N}(\mu, \Sigma)$.  The resulting distribution from rotating by $\pi/4$ and projecting onto the primary axis is $\mathcal{N} \left( [R \mu]_{1}, [R \Sigma R^{T}]_{11} \right)$ where
\begin{align}
[R \mu]_{1} &= \frac{\sqrt{2}}{2} ( \squaredMMDu{X_m}{Y_n}-\squaredMMDu{X_m}{Z_r} ) \\
[R \Sigma R^{T}]_{11}&= \frac{1}{2}(\sigma_{XY}^2 + \sigma_{XZ}^2 - 2 \sigma_{XYXZ}) \label{eq:denominator_of_the_TestStatistics}
\end{align}
with $R$ is the rotation  by $\pi/4.$  Then, the $p$-values for testing $\mathcal{H}_0$ versus $\mathcal{H}_1$ are
\begin{equation}
p \leq \Phi \left( -\frac{\squaredMMDu{X_m}{Y_n}-\squaredMMDu{X_m}{Z_r}}{\sqrt{\sigma_{XY}^2 + \sigma_{XZ}^2 - 2 \sigma_{XYXZ}}} \right)
\end{equation}
where $\Phi$ is the CDF of a standard normal distribution. We have made code for performing the test is available.\footnote{Code and examples can be found at \url{https://github.com/eugenium/MMD}}

%% file: sec_experiments.tex
\section{Experimental validation of the relative MMD test}
\label{sec:ExperimentsBasic}

We verify the validity of the hypothesis test described above using a synthetic data set in which we can directly control the relative similarity between distributions. 
%
%
%
We constructed three Gaussian distributions as illustrated in Figure~\ref{fig:synthetic_experiments}.  These Gaussian distributions are specified with different means so that we can control the degree of relative  similarity between them.  The question is whether the similarity between $X$ and $Z$ is greater than the similarity between $X$ and $Y$.  In these experiments, we used a Gaussian kernel with bandwidth selected as the median pairwise distance between data points, and we fixed $\mu_Y = [-20,-20]$, $\mu_Z = [20,20]$ and varied $\mu_X$ such that $\mu_X = ( 1-\gamma)\mu_Y + \gamma \mu_Z$, for 41 regularly spaced values of $\gamma\in [0.1,\; 0.9]$ (avoiding the degenerate cases $P_x=P_y$ or $P_x=P_z$). 
Figure~\ref{fig:synthetic_experiments_pvalues} shows the $p$-values of the relative similarity test for different distribution.  When $\gamma$ is varying around $0.5$, i.e., when $\squaredMMDu{X}{Y}$ is almost equal to $\squaredMMDu{X}{Z}$, the $p$-values quickly transition from $1$ to $0$, indicating strong discrimination of the test.  In Figure~\ref{fig:powerofthetest}, we compare the power of our test to the power of a naive test  when the reference sample is split in two, and the MMDs have no covariance: clearly, the latter simple approach does worse than ours  (a similar comparison in testing relative dependence returned the same advantage for a test based on the joint distribution; see \citet[Section 3]{bounliphone2015}).  Figure~\ref{fig:isocurve_conservativetest} shows an empirical scatter plot of the pairs of MMD statistics along with a $2\sigma$ iso-curve of the estimated distribution, demonstrating that the parametric Gaussian distribution is well calibrated to the empirical values. Futhermore, we validate our derived formulas using simulations in Appendix~\ref{sec:appendixB}, where we show the p-values have the correct distribution under the null.

\begin{minipage}{.45\textwidth}\centering
\includegraphics[width=0.65\textwidth]{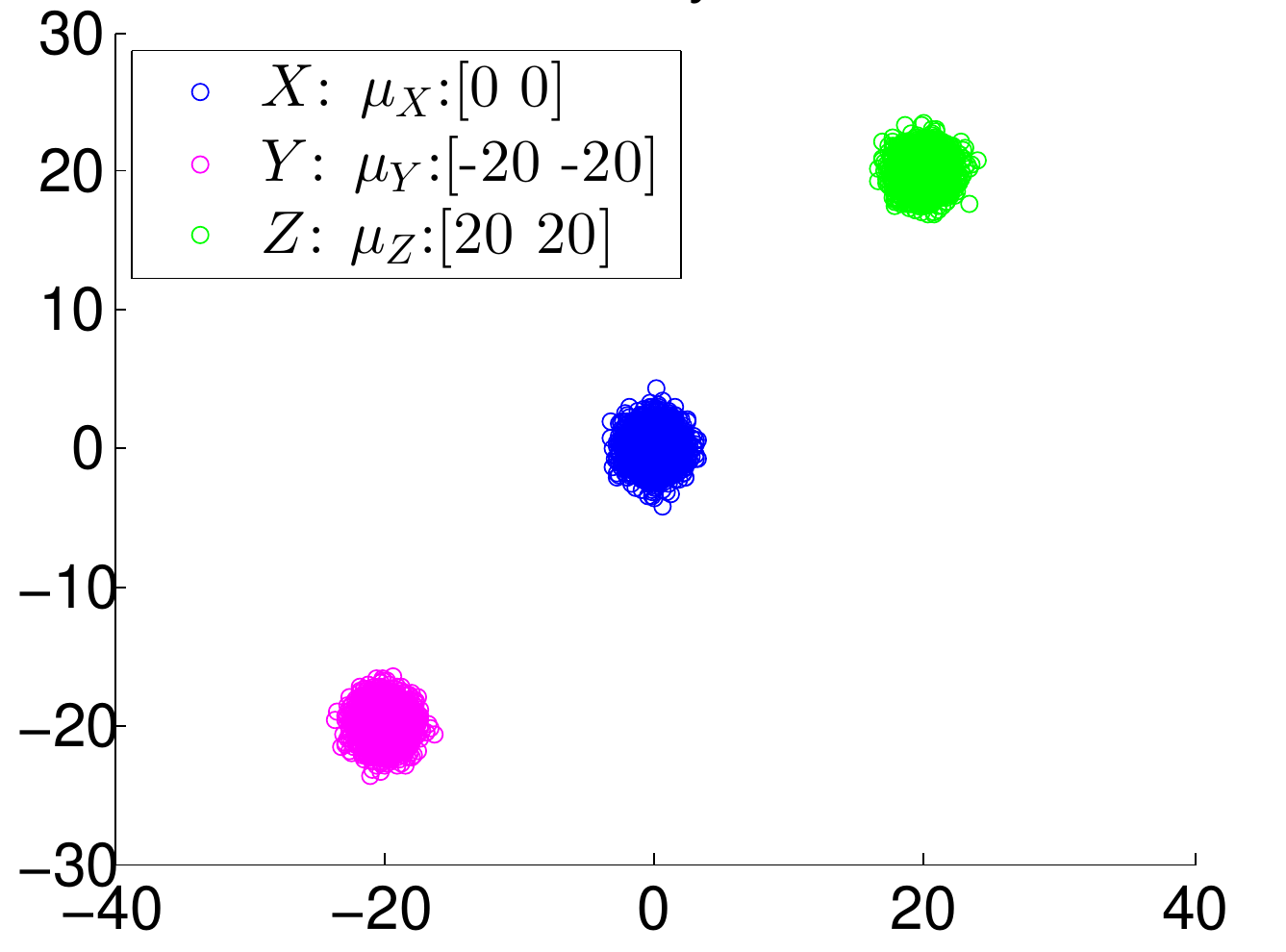}
\captionof{figure}{Illustration of the synthetic dataset where $X$, $Y$ and $Z$ are respectively Gaussian distributed with mean $\mu_X = [0,0]^T$, $\mu_Y=[-20,-20]^T$, $\mu_Z=[20,20]^T$ and with variance $\left(\protect\begin{smallmatrix}1&0\\0&1\protect\end{smallmatrix}\right)$.}
\label{fig:synthetic_experiments}
\end{minipage}
\hfill
\begin{minipage}{.45\textwidth}\centering
\begin{tabular}{cc}
\begin{sideways} $\qquad$ Power of the tests  \end{sideways} & \includegraphics[width=0.65\textwidth]{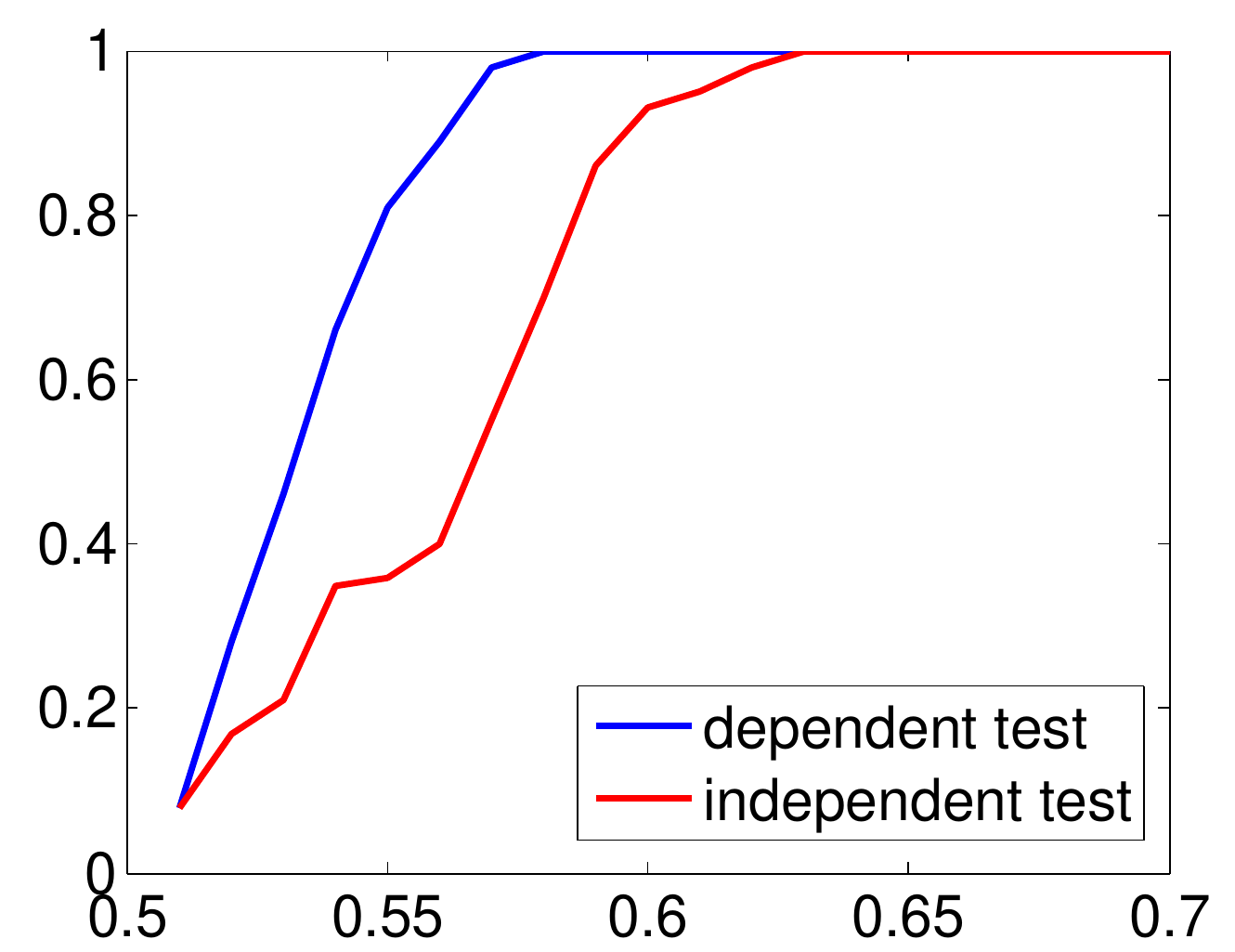} \\
& $\gamma$
\end{tabular}
\captionof{figure}{Comparison of the power of the proposed method to an independent test analogous to~\citet[Section 3]{bounliphone2015} as a function
of $\gamma$.
}
\label{fig:powerofthetest}
\end{minipage}

\begin{minipage}{\textwidth}
  \begin{minipage}[b]{0.49\textwidth}
    \centering
    \begin{tabular}{cc}
     \begin{sideways} $\qquad  \qquad$ $p$-values \end{sideways} & \includegraphics[width=.65\textwidth]{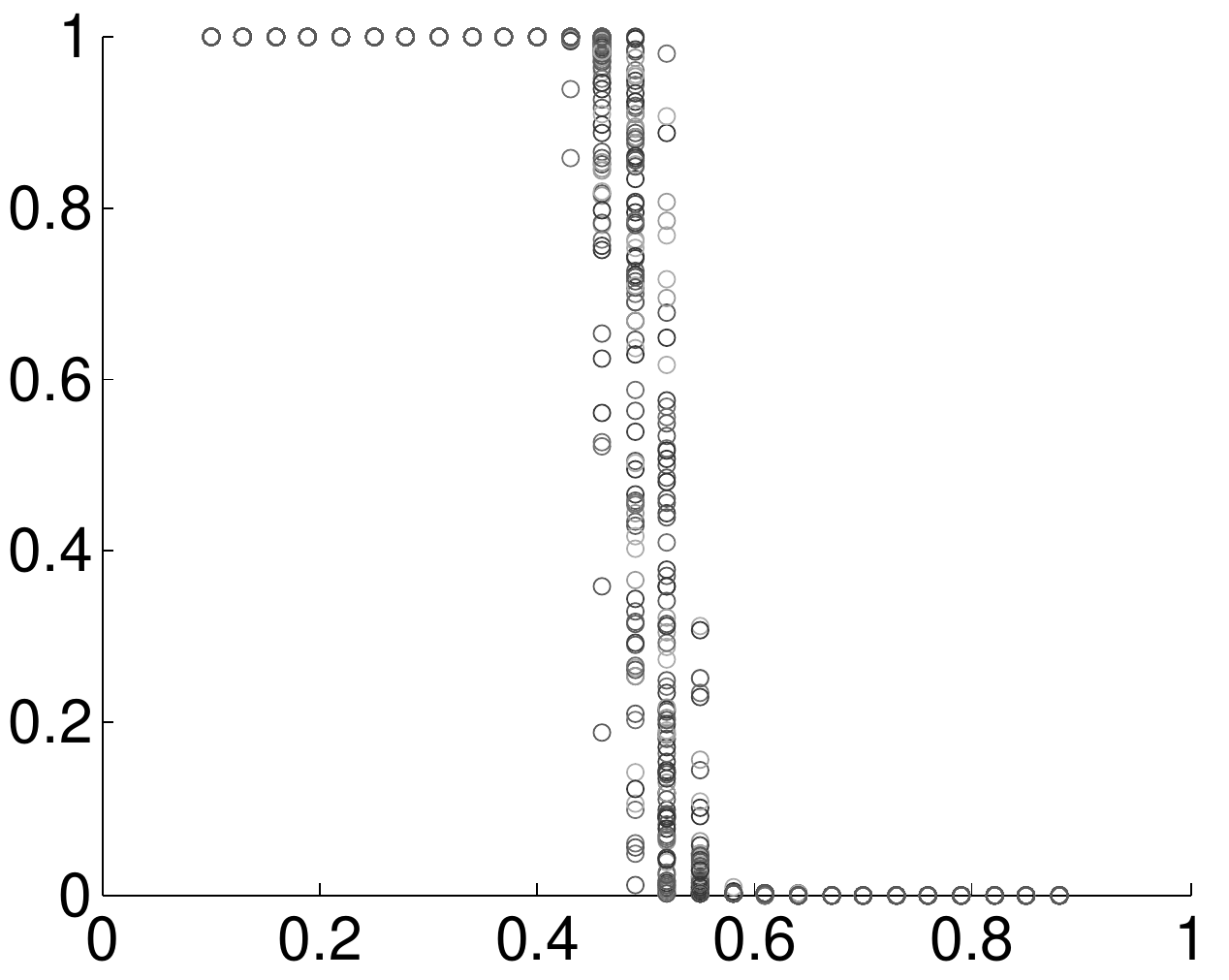} \\
     & $\gamma$ \\
     & $m=1000$ \\
     \end{tabular}
    \captionof{figure}{We fixed $\mu_Y = [-5,-5]$, $\mu_Z = [5,5]$ and varied $\mu_X$ such that $\mu_X = ( 1-\gamma)\mu_Y + \gamma \mu_Z$, for 41 regularly spaced values of $\gamma \in [0.1,\; 0.9]$ versus p-values for 100 repeated tests.}\label{fig:synthetic_experiments_pvalues}
  \end{minipage}
  \hfill
    \begin{minipage}[b]{0.49\textwidth}
    \centering
     \includegraphics[width=.65\textwidth]{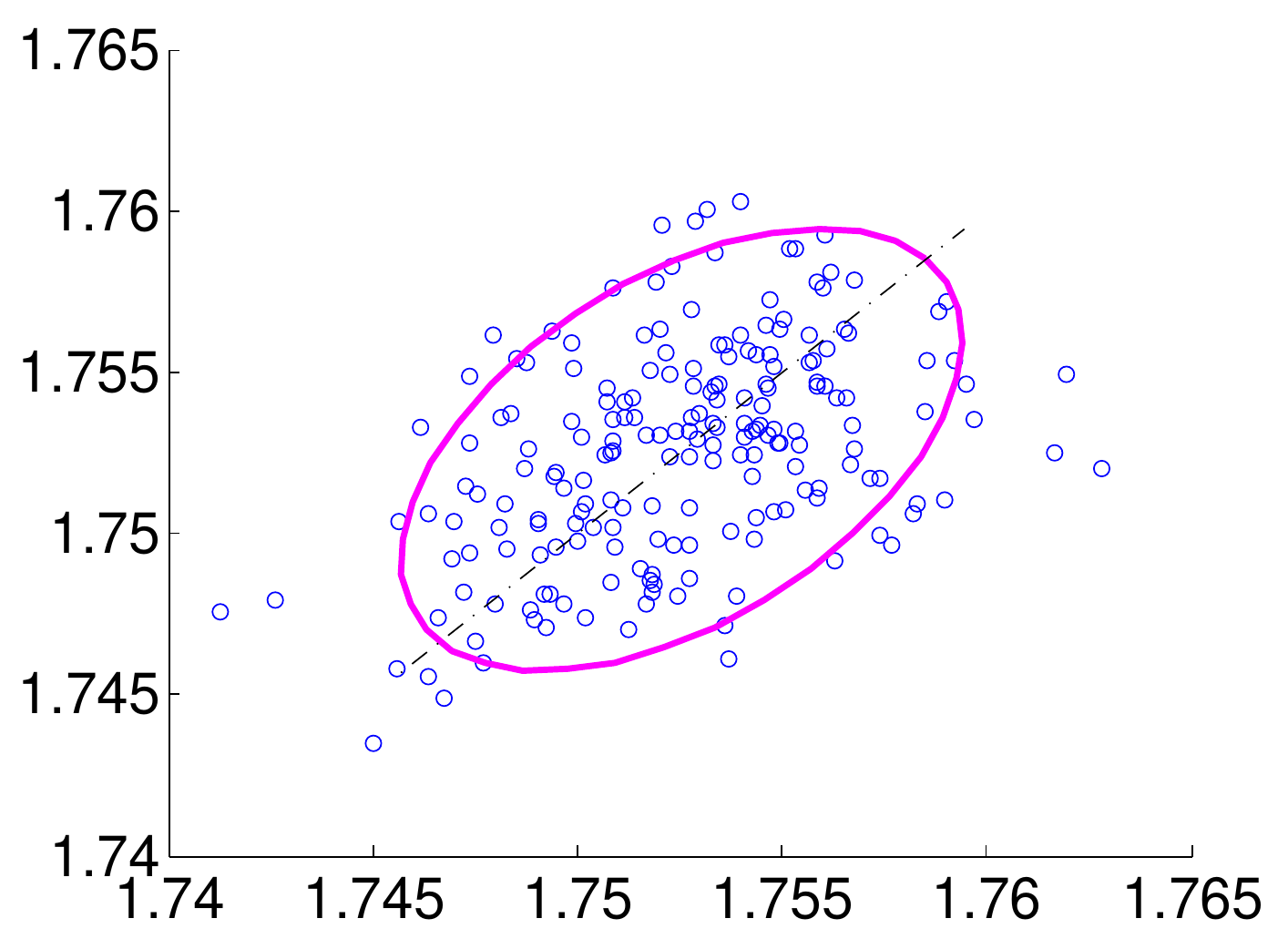} 
    \captionof{figure}{The empirical scatter plot of the joint MMD statistics with $m=1000$ for 200 repeated tests, along with the $2\sigma$ iso-curve of the analytical Gaussian distribution estimated by Equation~\eqref{eq:joint_asymptotic_MMD}.  The analytical distribution closely matches the empirical scatter plot, verifying the correctness of the variances.}\label{fig:isocurve_conservativetest}
  \end{minipage}
  \end{minipage}

%% file: sec_model_selection.tex
\input{subsec_experiments_autoencoder1.tex}

%% file: subsec_experiments_autoencoder1.tex
\section{Model Selection for Deep Unsupervised Neural Networks}
\label{sec:ModelSelectionDeepUnsupNN}
An important potential application of the Relative MMD can be found in recent work on unsupervised learning with deep neural networks \citep{kingma2013auto,bengio2013deep,larochelle2011neural,salakhutdinov2009deep,li2015generative,goodfellow2014generative}.  As noted by several authors, the evaluation of generative models is a challenging open problem \citep{li2015generative,goodfellow2014generative}, and the distributions of samples from these models are very complex and difficult to evaluate. The relative MMD performance can be used  to compare different model settings, or even model families, in a statistically valid framework. To compare two models using our test, we generate samples from both, and compare these to a set of real  target data samples that were not used to train either model. 

\begin{figure}\centering
\begin{tabular}{cc}
\includegraphics[width=0.28\textwidth]{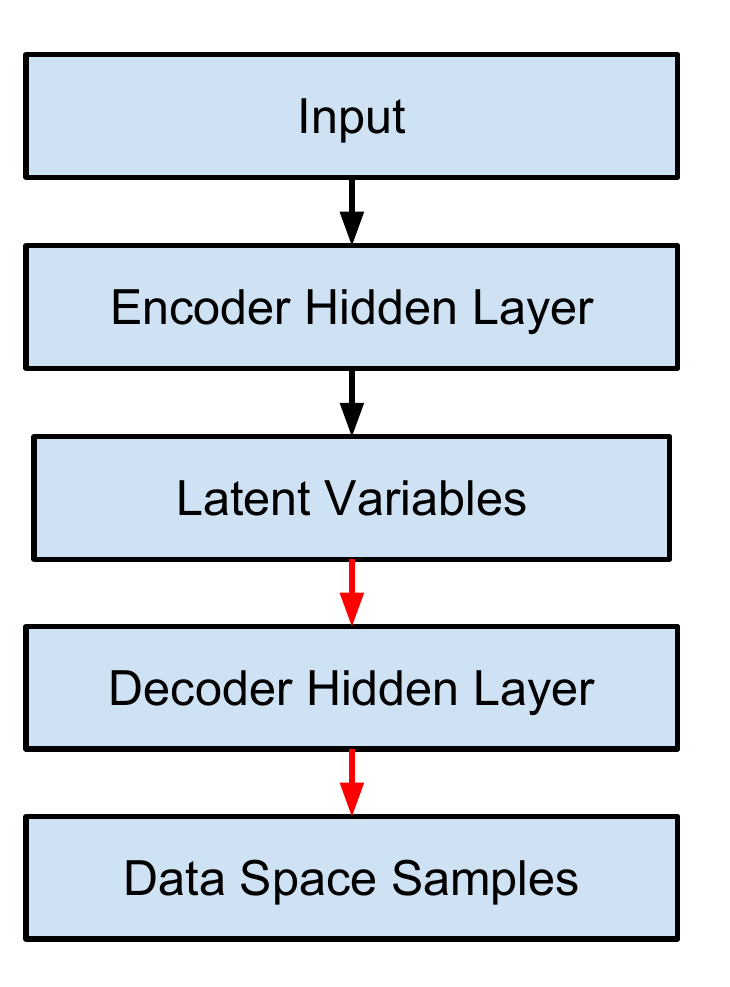} & 
\includegraphics[width=0.35\textwidth]{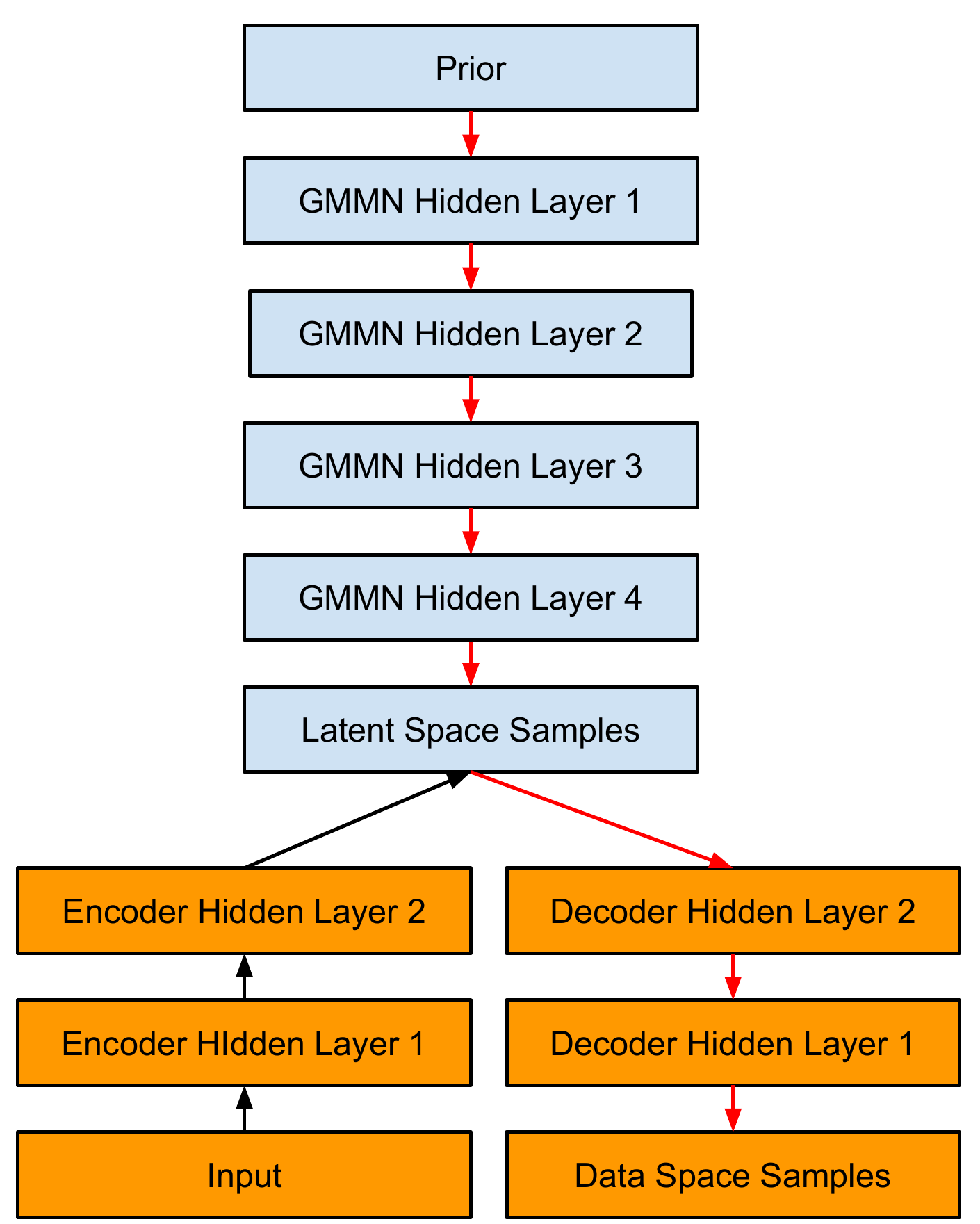} \\
(a) & (b) \\
\end{tabular}
\caption{(a) Variational auto-encoder reference model. We have 400 hidden nodes (both encoder and decoder) and 20 latent variables in the reference model for our experiments. (b) Auto-Encoder + GMMN reference model. The auto-encoder (indicated in orange) is trained separately and has 1024 and 32 hidden nodes in decode and encode hidden layers. The GMMN has 10 variables generated by the prior, and the hidden layers have 64, 256, 256, 1024 nodes in each layer respectively. In both networks red arrows indicate the data flow during sampling}
\label{fig:nets}
\end{figure}

In the experiments in the sequel we focus on the recently introduced variational auto-encoder (VAE) \citep{kingma2013auto} and the generative moment matching networks (GMMN)  \citep{li2015generative}. The former trains an encoder and decoder network jointly minimizing a regularized variational lower bound \citep{kingma2013auto}. While the latter class of models is purely generative minimizing an MMD based objective, this model works best when coupled with a separate auto-encoder which reduces the dimensionality of the data.  An architectural schematic for both classes of models is provided in Fig.~\ref{fig:nets}. Both these models can be trained using standard backpropagation \citep{Rumelhart:1988:LRB:65669.104451}. Using the latent variable prior we can directly sample the data distribution of these models without using MCMC procedures \citep{hinton2006fast,salakhutdinov2009deep}.  

We use the MNIST and FreyFace datasets for our analysis \citep{lecun1998gradient,kingma2013auto,goodfellow2014generative}. We first demonstrate the effectiveness of our test in a setting where we have a theoretical basis for expecting superiority of one unsupervised model versus another. Specifically, we use a setup where more training samples were used to create one model versus the other. We find that the Relative MMD framework agrees with the expected results (models trained with more data generalize better). We then demonstrate how the Relative MMD can be used in evaluating network architecture choices, and we show that our test strongly agrees with other established metrics, but in contrast can provide significance results using just the validation data while other methods may require an additional test set.  

Several practical matters must be considered when applying the Relative MMD test. The selection of kernel can affect the quality of results, particularly more suitable kernels can give a faster convergence. In this work we extend the logic of the median heuristic \citep{NIPS2012_4727} for bandwidth selection by computing the median pairwise distance between samples from $P_x$ and $P_y$ and averaging that with the median pairwise distance between samples from $P_x$ and $P_z$, which helps to maximize the difference between the two MMD statistics.
Although the derivations for the variance of our statistic hold for all cases, the estimates require asymptotic arguments and thus a sufficiently large $n$.  Selecting the kernel bandwidth in an appropriate range can therefore substantially increase the power of the test at a fixed sample size.  While we observed the median heuristic to work well in our experiments, there are cases where alternative choices of kernel can provide greater power: for instance, the kernel can be chosen to maximize the expected test power on a held-out dataset \citep{NIPS2012_4727}.  

\subsection{Variational Auto-Encoder Sample Size and Architecture Experiments}    
\begin{figure}[h]
    \centering
    \begin{subfigure}{.5\textwidth}
        \centering
        \includegraphics[width=1\textwidth]{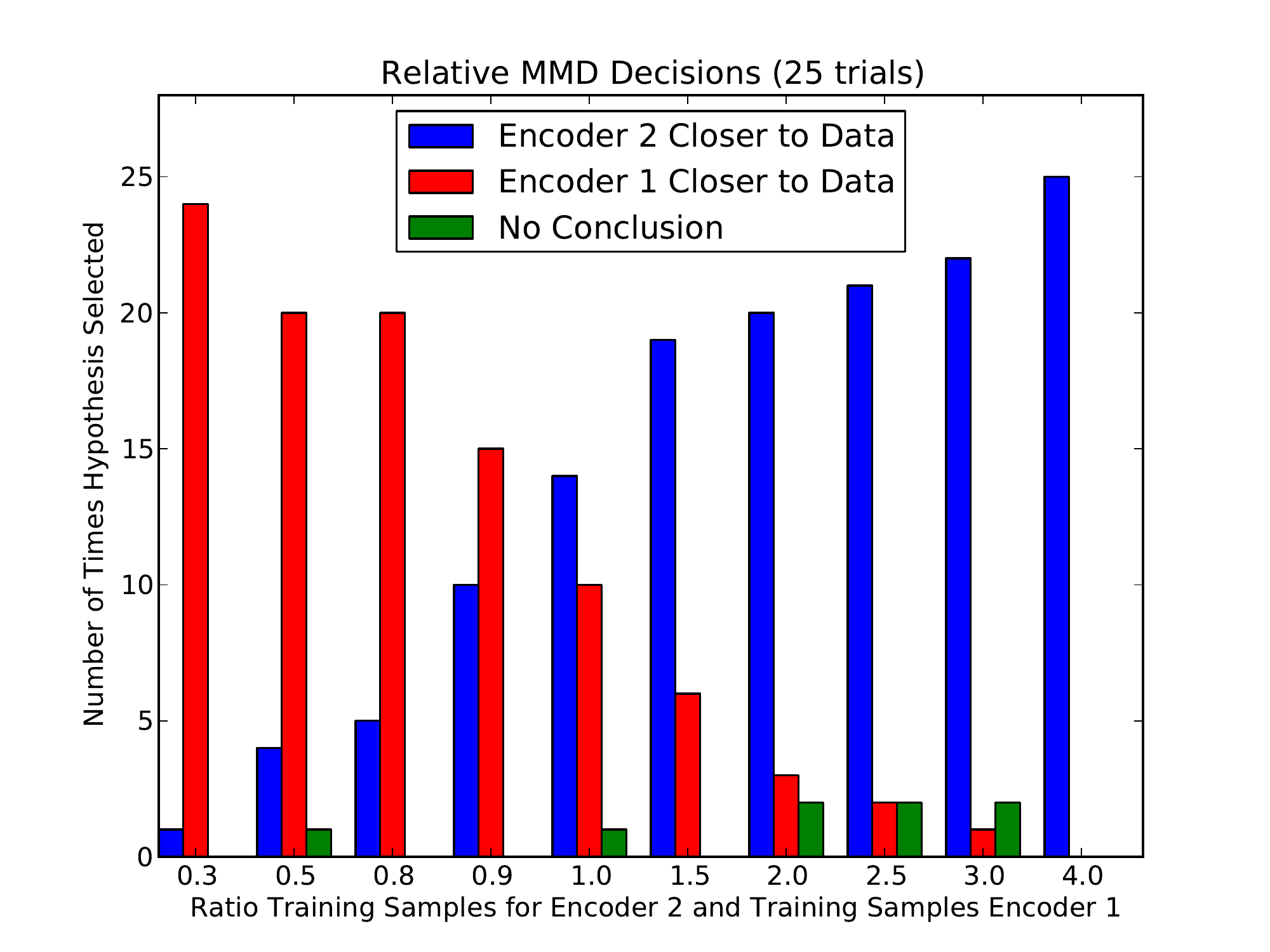}
        \caption{}
        \label{fig:sampsExp}
    \end{subfigure}\hfill
    \begin{subfigure}{.5\textwidth}
        \centering
        \includegraphics[width=1\textwidth]{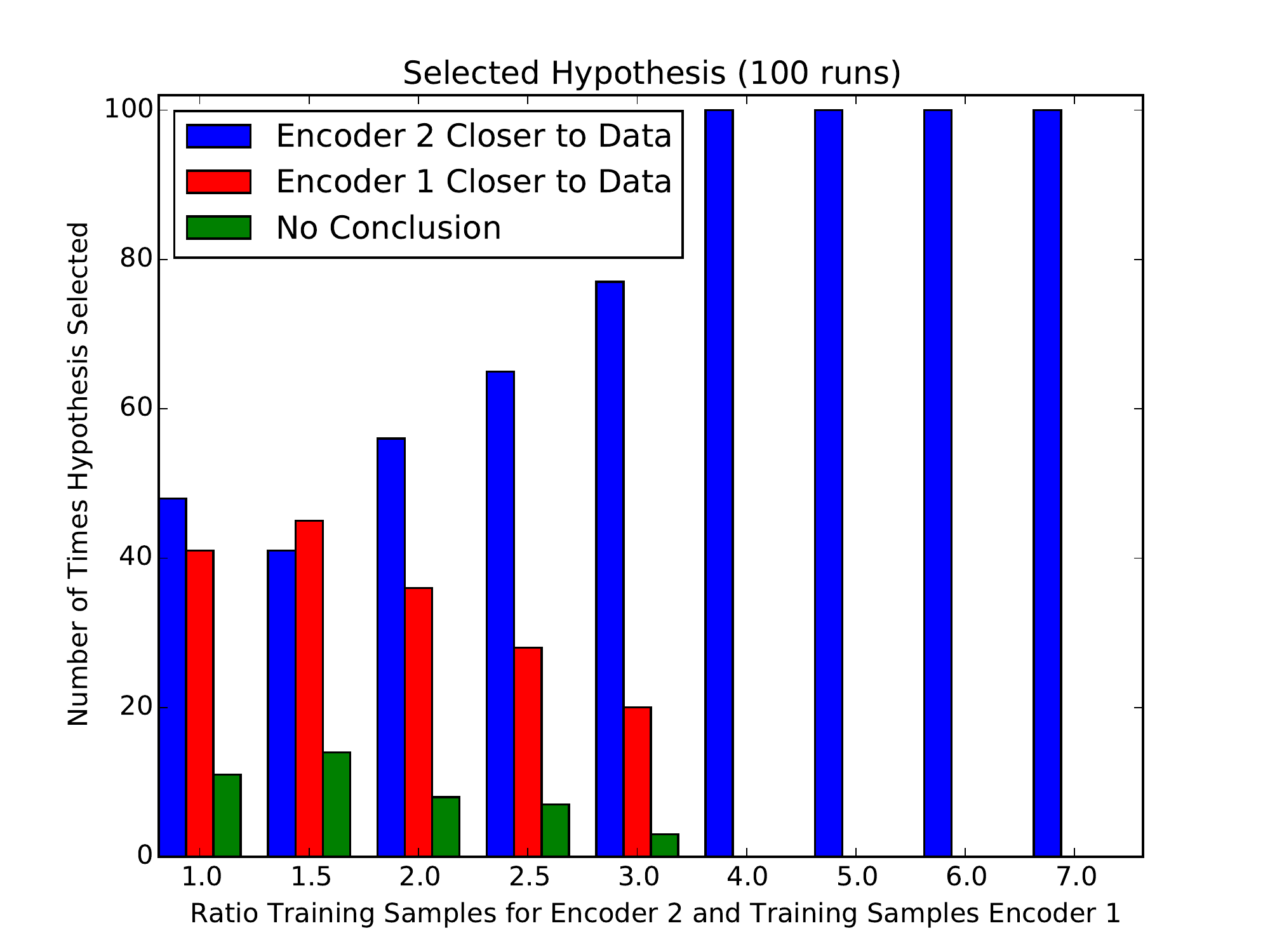}
        \caption{}
        \label{fig:sampsExp2}
    \end{subfigure}\medbreak
    \begin{subfigure}{.5\textwidth}
        \centering
        \includegraphics[width=1\textwidth]{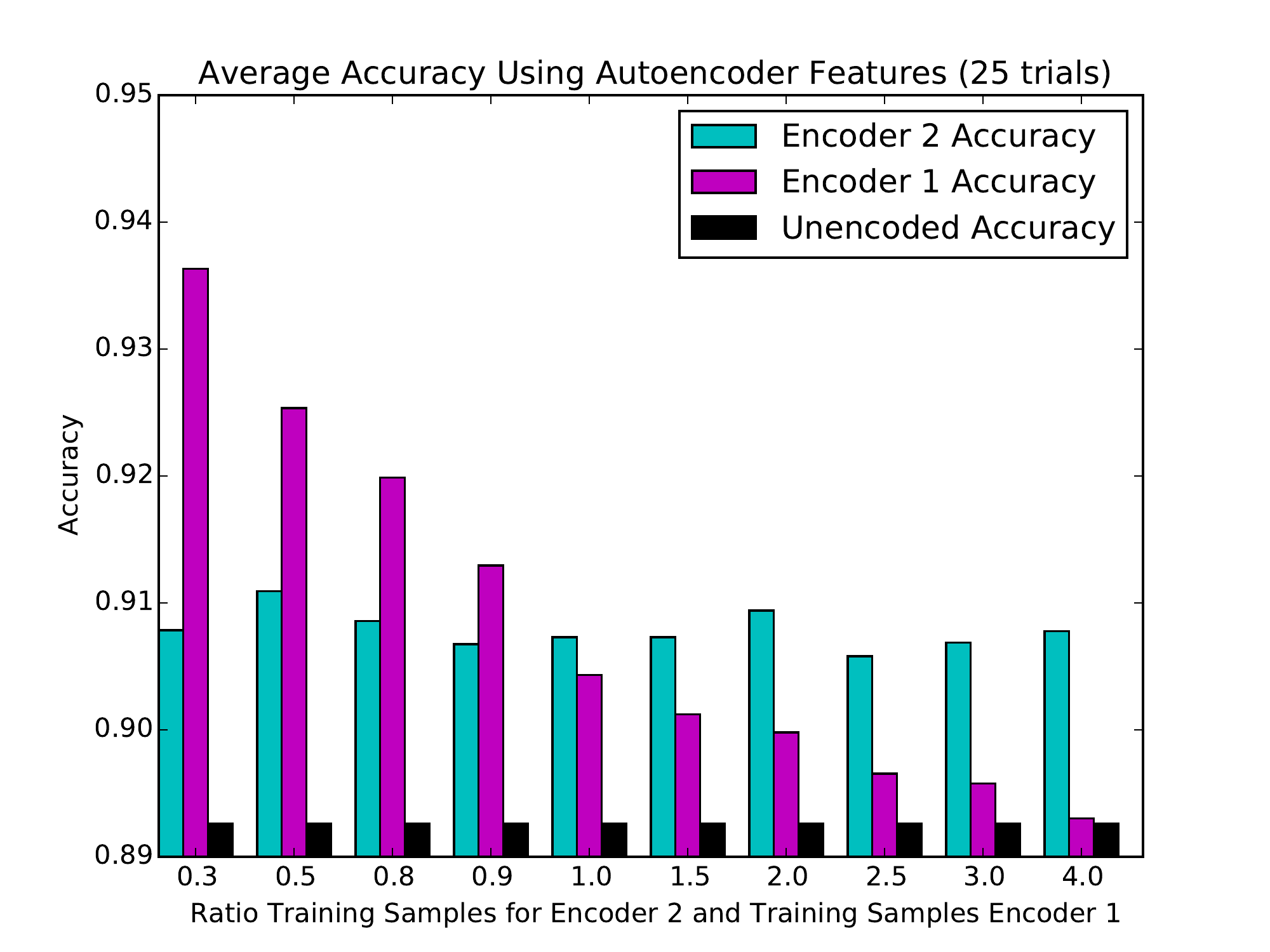}
        \caption{}
        \label{fig:sampsExp1}
    \end{subfigure}\hfill
    \begin{minipage}{.47\linewidth}
    \caption{We show the effect of (a) varying the training set size of one auto-encoder trained on MNIST data. (c) As a secondary validation we compute the classification accuracy of MNIST on a separate train/test set encoded using encoder 1 and encoder 2. (b) We then show the effect of varying the training set size of one auto-encoder using the FreyFace data. We note that due to the size of the FreyFace dataset, we limit the range of ratios used. From this figure we see that the results of the relative similarity test match our expectation: more data produces models which more closely match the true distribution.}
    \label{fig:prad3}
\end{minipage}

\end{figure}

We use the architecture from \cite{kingma2013auto} with a hidden layer at both the encoder and decoder and a latent variable layer as shown in Figure \ref{fig:nets}a. We use sigmoidal activation for the hidden layers of encoder and decoder. For the FreyFace data, we use a Gaussian prior on the latent space and data space. For MNIST, we used  a Bernoulli prior for the data space. 
 We fix the training set size of the second auto-encoder to 300 images for the FreyFace data and 1500 images for the MNIST data. We vary the number of training samples for the first auto-encoder. We then generate samples from both auto-encoders and compare them using Relative MMD to a held out set of data. We use 1500 FreyFace samples as the target in Relative MMD and 15000 images from MNIST. Since a single sample of the data might lead to better generalization performance by chance, we repeat this experiment multiple times and record whether the relative similarity test indicated a network is preferred or if it failed to reject the null hypothesis. The results are shown in \autoref{fig:prad3} which demonstrates that we are closely following the expected model preferences. Additionally for MNIST we use another separate set of supervised training and test data. We encode this data using both auto-encoders and use logistic regression to obtain a classification accuracy. The indicated accuracies closely match the results of the relative similarity test, further validating the test.

We consider model selection between networks using different architectures.  We train two encoders, one a fixed reference model (400 hidden units and 20 latent variables), and the other varying as specified in Table~\ref{tab:VAEarch}. 25000 images from the MNIST data set were used for training. We use another 20000 images as the target data in Relative MMD. Finally, we use a set of 10000 training and 10000 test images for a  supervised task experiment. We use the labels in the MNIST data and perform training and classification using an $\ell_2$-regularized logistic regression on the encoded features. 
In addition we use the supervised task test data to evaluate the variational lower bound of the data under the two models \citep{kingma2013auto}. We show the result of this experiment in Table \ref{tab:VAEarch}. For each comparison we take a different subset of training data which helps demonstrate the variation in lower bound and accuracy when re-training the reference architecture. We use a significance value of $5\%$ and indicate when the test favors one auto-encoder over another or fails to reject the null hypothesis. We find that Relative MMD evaluation of the models closely matches performance on the supervised task and the test set variational lower bound.
%
%
%

\begin{table}[h]\centering
\resizebox{1\textwidth}{!}{
\begin{tabular}{l|l|l|l|l|l|l}
Hidden&Latent  &  Result   & Accuracy (\%) & Accuracy (\%) & Lower Bound & Lower Bound  \\
 VAE 1  & VAE 1& RelativeMMD & VAE 1 &VAE 2 & VAE 1 &VAE 2 \\\hline
200 & 5 & Favor VAE 2 & $92.8\pm 0.3$      & $\bm{94.7 \pm 0.2}$ & -126  &\textbf{-97}\\
200 & 20 & Favor VAE 2  & $92.6\pm 0.3$      & $\bm{94.5 \pm 0.2}$ &-115 &\textbf{-105}    \\
400 & 50 & Favor VAE 1  & $\bm{94.6\pm 0.2}$      & $94.0\pm 0.2$ & \textbf{-99.6} &-123.44\\
800 & 20 & Favor VAE 1  & $\bm{94.8 \pm 0.2}$      & $93.9\pm 0.2$ & \textbf{-111} &-115\\
800 & 50 & Favor VAE 1  & $94.2\pm 0.3$      & $94.5\pm 0.2$ &\textbf{-101} &-103
\end{tabular}
}
\caption{We compare several variational auto encoder (VAE) architectural choices for the number of hidden units in both decoder and encoder and the number of latent variables for the VAE. The reference encoder, denoted encoder 2, has 400 hidden units and 20 latent variables. We denote the competing architectural models as encoder 1. 
We vary the number of hidden nodes in both the decoder and encoder and the number of latent variables. 
Our test closely follows the performance difference of the auto-encoder on a supervised task (MNIST digit classification) as well as the variational lower bound on a withheld set of data. The data used for evaluating the Accuracy and Lower Bound is separate from that used to train the auto-encoders and for the hypothesis test.} 
\label{tab:VAEarch}
\end{table}

\subsection{Generative Moment Matching Networks Architecture Experiments}
We demonstrate our hypothesis test on a different class of deep generative models called Generative Moment Matching Networks (GMMN) \citet{li2015generative}. This recently introduced model has shown competitive performance in terms of test set likelihood on the MNIST data. Furthermore the training of this model is based on the MMD criterion. \cite{li2015generative} proposes to use that model along with an auto-encoder, which is the setup we employ in this work. Here a standard auto-encoder model is trained on the data to  obtain a low dimensional representation, then a GMMN network is trained on the latent representations (Figure~\ref{fig:nets}).  

We use the relative similarity test  to evaluate various architectural choices in this new class of models. We start from the baseline model specified in \cite{li2015generative} and associated software. The details of the reference model are specified in Figure~\ref{fig:nets}.

We vary the number of auto-encoder hidden layers (1 to 4), generative model layers(1, 4, or 5), the number of network nodes (all or 50\% of the reference model), and use of drop-out on the auto-encoder. We use the same training set of 55000, validation set of 5000 and test set of 10000 as in \citep{li2015generative,goodfellow2014generative}. In total we train 48 models. We use these to compare 4 simplified binary network architecture choices using the Relative MMD: using dropout on the auto-encoder, few (1) or more (4 or 5) GMMN layers, few (1 or 2) or more (3 or 4) auto-encoder layers, and the number of network nodes. We use our test to compare these model settings using the \emph{validation set} as the target in the relative similarity test, and samples from the models as the two sources. To validate our results we compare it to likelihoods computed on the test set.  The results are shown in Table~\ref{tab:mmd}. We see that the likelihood results computed on a separate test set follow the conclusions obtained from MMD on the validation set. Particularly, we find that using fewer hidden layers for the GMMN and more hidden nodes generally produces better models.
 
\begin{table}[h]
\resizebox{1\textwidth}{!}{
\begin{tabular}{l|c|c|c|c|c}
& \multicolumn{3}{c|}{RelativeMMD Preference}   &  \multicolumn{2}{c}{}  \\\hline
Experimental Condition (A/B)& A & Inconclusive 
& B & Avg Likelihood A  & Avg Likelihood B \\\hline
Dropout/No Dropout&199                  & 17         & 360                    & $-9.01    \pm 55.43$  & $76.76 \pm 42.83$          \\\hline
More/Fewer GMMN Layers&105                     & 14          & 393                    & $-73.99 \pm 40.96  $   & $\bm{249.6 \pm 8.07 }$           \\\hline
More/Fewer Nodes&450                    & 13          & 113                     & $\bm{125.2   \pm 43.4}$    & $-57 \pm 49.57$    \\\hline
More/Fewer AE layers&231                  & 21 & 324                    & $41.78 \pm 44.07 $       & $25.96 \pm 55.85$              
\end{tabular}
}
\caption{For each experimental condition (e.g. dropout or no dropout) we show the number of times the Relative MMD prefers models in group 1 or 2 and number of inconclusive tests. We use the validation set as the target data for Relative MMD. An average likelihood for the MNIST test set for each group is shown with error bars. We can see that the MMD choices are in agreement with likelihood evaluations. Particularly we identify that models with fewer GMMN layers  and models with more nodes have more favourable samples, which is confirmed by the likelihood results.}
\label{tab:mmd}
\end{table}

%% file: AutoEncExpDiscussion.tex
\subsection{Discussion}
In these experiments we have seen that the Relative MMD test can be used to compare deep generative models obtaining judgements aligned with other metrics. Comparisons to other metrics are important for verifying our test is sensible, but it can occlude the fact that MMD is a valid evaluation technique on its own. When evaluating only sample generating models where likelihood computation is not possible, MMD is an appropriate and tractable metric to consider in addition to Parzen-Window log likelihoods and visual appearance of the samples. In several ways it is potentially more appropriate than Parzen-windows as it allows one to consider directly the discrepancy between the test data samples and the model samples while allowing for significance results. In such a situation, comparing the performance of several models using the MMD against a single set of test samples, the Relative MMD test can provide an automatic significance value without expensive cross-validation procedures.  

Gaussian kernels are closely related to Parzen-window estimates, thus computing an MMD in this case can be considered related to comparing Parzen window log-likelihoods. The MMD gives several advantages, however. 
First, the asymptotics of MMD are quite different to Parzen-windows, since the Parzen-window bandwidth shrinks as $n$ grows. Asymptotics of relative tests with shrinking bandwidth are unknown: even for two samples this is challenging \citep{KrishnamurthyKP15}. Other two sample tests are not easily extendable to relative tests~\citep{rosenbaum2005exact, friedman1979multivariate,hall2002permutation}. This is because  the tests above rely on graph edge counting or nearest neighbor-type statistics, and null distributions are obtained via combinatorial arguments which are not easily extended from two to three samples. MMD is a $U$-statistic, hence its asymptotic behavior is much more easily generalised to multiple dependent statistics.

There are two primary advantages of the MMD over the variational lower bound, where it is known \citep{kingma2013auto}: first, we have a characterization of the asymptotic behavior, which allows us to determine when the difference in performance is significant; second, comparing two lower bounds produced from two different models is unreliable, as we do not know how conservative either lower bound is.


%% file: conclusion.tex
\section{Conclusion}

We have described a novel non-parametric statistical hypothesis test for relative similarity based on the Maximum Mean Discrepancy. The test is consistent, and the computation time is quadratic.  Our proposed test statistic is theoretically justified for the task of comparing samples from arbitrary distributions as it can be shown to converge to a quantity which compares all moments of the two pairs of distributions.



We evaluate test performance on  synthetic data, where the degree of similarity can be controlled. 
Our experimental results on model selection for deep generative networks show that Relative MMD can be a useful approach to comparing such models. There is a strong correspondence between the test resuts and the expected likelihood, prediction accuracy, and variational lower bounds on the models tested. Moreover, our test  has the advantage over these alternatives of providing guarantees of statistical significance to its conclusions.
This suggests that the relative similarity test will be useful in evaluating hypotheses about network architectures, for example that AE-GMMN models may generalize better when fewer layers are used in the generative model. 
Code for our method is available.\footnote{ \url{https://github.com/eugenium/MMD}}


%% file: detailedCovarianceDerivations.tex
\section{Detailed Derivations of the Test Variance and Covariance}\label{sec:appendixA}

The variance and the covariance for a $U$-statistic is described in \citet[Eq. 5.13]{hoeffding1948class} and \citet[Chap. 5]{serfling2009approximation}.

Let $\mathcal{V} := (v_1, ..., v_m)$ be $m$ iid random variables where $v:= (x,y) \sim P_x \times P_y$. An unbiased estimator of $\squaredMMDpop{x}{y}$ is 
\begin{equation}
\squaredMMDu{X_m}{Y_n} = \frac{1}{m(m-1)} \sum_{i \ne j}^m h(v_i,v_j)
\end{equation}
with $h(v_i,v_j) = k(x_i,x_j) + k(y_i,y_j) -k(x_i,y_j)  -k(x_j,y_i)$.

Similarly, let $\mathcal{W} := (w_1, ..., w_m)$ be $m$ iid random variables where $w:= (x,z) \sim P_x \times P_z$. An unbiased estimator of $\squaredMMDpop{x}{z}$ is 
\begin{equation}
\squaredMMDu{X_m}{Z_r} = \frac{1}{m(m-1)} \sum_{i \ne j}^m g(w_i,w_j)
\end{equation}
with $g(w_i,w_j) = k(x_i,x_j) + k(z_i,z_j) -k(x_i,z_j)  -k(x_j,z_i)$

Then the variance/covariance for a $U$-statistic with a kernel of order 2 is given by
\begin{align}
Var(\operatorname{MMD}_u^2) &= \frac{4(m-2)}{m(m-1)} \zeta_1 + \frac{2}{m(m-1)} \zeta_2 
\label{eq:allterm_in_variance_of_MMDstatistics}
\end{align}
Equation~\eqref{eq:allterm_in_variance_of_MMDstatistics} with neglecting higher terms can be written as
\begin{align}
Var(\operatorname{MMD}_u^2) & = \frac{4(m-2)}{m(m-1)} \zeta_1 + \mathcal{O}
(m^{-2}) 
\label{eq:derivation_of_the_variance_ofMMD}
\end{align}
where for the variance term, $\zeta_1 = \var \left[ \Eone \left[ h(v_1,V_2) \right] \right] $ and for the covariance term $ \zeta_1 = \var \left[ \operatorname{\E}_{v_1,w_1} \left[ h(v_1,V_2)g(w_1,W_2) \right] \right]$.

\paragraph{Notation} $[\tilde{K}_{xx}]_{ij} = [K_{xx}]_{ij}$ for all $i \ne j$ and $[\tilde{K}_{xx'}]_{ij} =0$ for $j =i$. Same for $\tilde{K}_{yy}$ and $\tilde{K}_{zz}$. We will also make use of the fact that $k(x_i,x_j) = \langle \phi(x_i), \phi(x_j) \rangle$ for an appropriately chosen inner product, and  function $\phi$.  We then denote
\begin{equation}
  \mu_x := \int \phi(x) dP_x .
\end{equation}

\subsection{Variance of MMD}

We note many terms in expansion of the squares above cancel out due to independence. For example  $\E_{x_1,y_1}\left[ \langle \phi(y_1), \mu_y \rangle \langle \phi(x_1),\mu_y \rangle \right] - \E_{y_1} \left[ \langle \phi(y_1), \mu_y \rangle \right]  \E_{x_1}\left[ \langle \phi(x_1),\mu_y \rangle \right]=0$. 

We can thus simplify to the following expression for $\zeta_1$

\begin{align}
\zeta_1 &= \EXoneYone \left[ \left( \EXtwoYtwo \left[h(x_1,y_1)\right] \right) ^2 \right] - \left( \squaredMMDpop{X}{Y} \right)^2 \\
&= \EXoneYone \left[ (\langle \phi(x_1) , \mu_x \rangle + \langle \phi(y_1), \mu_y \rangle - \langle \phi(x_1), \mu_y \rangle - \langle \mu_x , \phi(y_1) \rangle)^2 \right] - \left( \squaredMMDpop{X}{Y} \right)^2 \\ 
&= \EXoneYone \big[ 
\langle \phi(x_1),\mu_x \rangle^2 
+2 \langle \phi(x_1),\mu_x \rangle \langle \phi(y_1),\mu_y \rangle 
-2 \langle \phi(x_1),\mu_x \rangle \langle \phi(x_1),\mu_y \rangle 
\\ & \nonumber
\qquad \qquad
-2 \langle \phi(x_1),\mu_x \rangle \langle \phi(y_1),\mu_x \rangle
+ \langle \phi(y_1), \mu_y \rangle^2
\\& \nonumber
\qquad \qquad
-2 \langle \phi(y_1), \mu_y \rangle \langle \phi(x_1),\mu_y \rangle
-2 \langle \phi(y_1), \mu_y \rangle \langle \phi(y_1), \mu_x \rangle
\\ & \nonumber
\qquad \qquad
+ \langle \phi(x_1),\mu_y \rangle^2
+2 \langle \phi(x_1),\mu_y \rangle \langle \phi(y_1),\mu_x \rangle 
\\& \nonumber
\qquad \qquad
+\langle \phi(y_1),\mu_x \rangle^2 
\big] - \left( \squaredMMDpop{X}{Y} \right)^2 \\
&= \label{eq:MMDvarPopulationTerms}
\EXone [\langle \phi(x_1),\mu_x \rangle^2] - \EXone [\langle \phi(x_1),\mu_x \rangle]^2
\\& \nonumber
\qquad \qquad
-2 (\EXone [\langle \phi(x_1),\mu_x \rangle \langle \phi(x_1),\mu_y \rangle] - \EXone [\langle \phi(x_1),\mu_x \rangle] \EXone [\langle \phi(x_1),\mu_y \rangle])
\\& \nonumber
\qquad \qquad
+ \EYone[\langle \phi(y_1), \mu_y \rangle^2] - \EYone [\langle \phi(y_1), \mu_y \rangle]^2
\\ & \nonumber
\qquad \qquad
-2 ( \EYone [\langle \phi(y_1), \mu_y \rangle \langle \phi(y_1), \mu_x \rangle ] - \EYone [\langle \phi(y_1), \mu_y \rangle] \mathbb{E}_{y_1} [ \langle \phi(y_1), \mu_x \rangle ])
\\ & \nonumber
\qquad \qquad
+ \EXone [\langle \phi(x_1),\mu_y \rangle^2] - \EXone [\langle \phi(x_1),\mu_y \rangle]^2
\\ & \nonumber
\qquad \qquad
+ \EYone [\langle \phi(y_1),\mu_x \rangle^2 ] - \EYone [\langle \phi(y_1),\mu_x \rangle]^2 
\end{align}

Substituting empirical expectations over the data sample for the population expectations in Eq.~\eqref{eq:MMDvarPopulationTerms} gives
\begin{align}
\zeta_1 & \approx
\frac{1}{m(m-1)^2} e^T \tilde{K}_{xx}\tilde{K}_{xx} e - \left(\frac{1}{m(m-1)} e^T \tilde{K}_{xx} e \right)^2
\\ & \nonumber
\qquad \qquad
-2 \left(
\frac{1}{m(m-1) n} e^T \tilde{K}_{xx} K_{xy} e
- \frac{1}{m^2(m-1)n} e^T \tilde{K}_{xx} e e^T K_{xy} e
\right)
\\ & \nonumber
\qquad \qquad
+ \frac{1}{n(n-1)^2} e^T \tilde{K}_{yy} \tilde{K}_{yy} e
- \left( \frac{1}{n(n-1)} e^T \tilde{K}_{yy} e \right)^2
\\ & \nonumber
\qquad \qquad
-2 \left( \frac{1}{n(n-1)m} e^T \tilde{K}_{yy} K_{yx} e
- \frac{1}{n^2(n-1)m} e^T \tilde{K}_{yy} e  e^T K_{xy} e
\right)
\\ & \nonumber
\qquad \qquad
+ \frac{1}{n^2 m} e^T K_{yx} K_{xy} e
- 2 \left( \frac{1}{nm} e^T K_{xy} e \right)^2
+ \frac{1}{m^2 n} e^T K_{xy} K_{yx} e   \nonumber
\end{align}

Derivation of the first term for example
\begin{align}
\EXone [\langle x_1,\mu_x \rangle^2] & \approx \frac{1}{m} \sum_{i=1}^m \langle \phi(x_i), \frac{1}{m-1}\sum_{j=1 \atop j \ne i}^m \phi(x_j) \rangle \langle \phi(x_i),\frac{1}{m-1}\sum_{k=1 \atop k \ne i}^m \phi(x_k) \rangle \\
&= \frac{1}{m(m-1)^2} \sum_{i=1}^m  \sum_{j=1 \atop j \ne i}^m \sum_{k=1 \atop k \ne i}^m k(x_i,x_j)k(x_i,x_k) \nonumber \\
&= \frac{1}{m(m-1)^2} e^T \tilde{K}_{xx} \tilde{K}_{xx} e \nonumber
\end{align}

\subsection{Covariance of MMD}

We note many terms in expansion of the squares above cancel out due to independence. For example  $\E_{x_1,z_1} \left[\langle \phi(x_1),\mu_x \rangle \langle \phi(z_1),\mu_z \rangle \right] - \E_{x_1} \left[ \langle \phi(x_1),\mu_x \rangle \right]  \E_{z_1}\left[ \langle \phi(z_1),\mu_z \rangle \right]=0$. 

We can thus simplify to the following expression for $\zeta_1$

\begin{align}
\zeta_1 &= \EXoneYoneZone \left[  \EXtwoYtwoZtwo \left[ h(x_1,y_1)  g(x_1,z_1) \right]\right] - \left( \squaredMMDpop{X}{Y} \squaredMMDpop{X}{Z} \right) \\
&= \EXoneYoneZone [ (\langle \phi(x_1),\mu_x \rangle + \langle \phi(y_1),\mu_y \rangle -\langle \phi(x_1),\mu_y \rangle - \langle \phi(x_1),\mu_y) \rangle) \nonumber \\ 
& \qquad \qquad \qquad  (\langle \phi(x_1),\mu_x) \rangle + \langle \phi(z_1),\mu_z \rangle -\langle \phi(x_1),\mu_z \rangle - \langle \phi(x_1),\mu_z \rangle) ] \nonumber \\
& \qquad \qquad  -\squaredMMDpop{X}{Y} \squaredMMDpop{X}{Z} \nonumber \\ 
&= \EXone \left[ \langle \phi(x_1),\mu_x \rangle^2 \right] - \EXone \left[ \langle \phi(x_1),\mu_x \rangle \right]^2 \\
& \qquad - \left( \EXone \left[ \langle \phi(x_1),\mu_x \rangle \langle \phi(x_1),\mu_z \rangle \right] - \EXone \left[ \langle \phi(x_1),\mu_x \rangle \right] \EXone \left[ \langle \phi(x_1),\mu_z \rangle \right] \right) \nonumber \\
&  \qquad - \left( \EXone \left[ \langle \phi(x_1),\mu_x \rangle \langle \phi(x_1),\mu_y \rangle \right] - \EXone \left[ \langle \phi(x_1),\mu_x \rangle \right] \EXone \left[ \langle \phi(x_1),\mu_y \rangle \right] \right) \nonumber \\
&  \qquad + \EXone \left[ \langle \phi(x_1),\mu_y \rangle \langle \phi(x_1),\mu_z \rangle \right] - \EXone \left[ \langle \phi(x_1),\mu_y \rangle \right] \EXone \left[  \langle \phi(x_1),\mu_z \rangle \right]  \nonumber \\
& \approx
\frac{1}{m(m-1)^2} e^T \tilde{K}_{xx}\tilde{K}_{xx} e - \left(\frac{1}{m(m-1)} e^T \tilde{K}_{xx} e \right)^2 \nonumber \\
& \qquad - \left( \frac{1}{m(m-1)r} e^T \tilde{K}_{xx} K_{xz} e - \frac{1}{m^2(m-1)r}  e^T \tilde{K}_{xx} e e^T K_{xz} e \right) \nonumber \\
& \qquad - \left( \frac{1}{m(m-1)n} e^T \tilde{K}_{xx} K_{xy} e - \frac{1}{m^2(m-1)n} e^T \tilde{K}_{xx} e e^T K_{xz} e \right) \nonumber \\
& \qquad + \left( \frac{1}{mnr} e^T K_{yx} K_{xz} e - \frac{1}{m^2nr} e^T K_{xy}e e^T K_{xz} e\right) \nonumber
\end{align}

\subsection{Derivation of the variance of the difference of two MMD statistics}
In this section we propose an alternate strategy of deriving directly the variance of a u-statistic of the difference of MMDs with a joint variable. This formulation agrees with the derivation of the covariance matrix and subsequent projection, and provides extra insights.

Let $\mathcal{D} := (d_1, ..., d_m)$ be $m$ iid random variables where $d:= (x,y,z) \sim P_x \times P_y \times P_z$.  Then the difference of the unbiased estimators of $\squaredMMDpop{x}{y}$ and $\squaredMMDpop{x}{z}$ is given by
\begin{equation}
\squaredMMDu{x}{y} - \squaredMMDu{x}{z} = \frac{1}{m(m-1)} \sum_{i \ne j}^m f(d_i,d_j)
\label{eq:diff_stat}
\end{equation}
with $f$, the kernel of $\squaredMMDpop{x}{y} - \squaredMMDpop{x}{z}$ of order 2 as follows
\begin{align}\nonumber
f(d_1,d_2) &= (k(x_1,x_2) + k(y_1,y_2) - k(x_1,y_2) - k(x_2,y_1)) 
\\ \nonumber&
\qquad - (
k(x_1,x_2) + k(z_1,z_2) - k(x_1,z_2) - k(x_2,z_1)
)
\\ &=
(k(y_1,y_2) - k(x_1,y_2) - k(x_2,y_1))
- (
k(z_1,z_2) - k(x_1,z_2) - k(x_2,z_1)
)
\end{align}

Equation \eqref{eq:diff_stat} is a $U$-statistic and thus we can apply Equation \eqref{eq:derivation_of_the_variance_ofMMD} to obtain its variance. We first note
\begin{align}\nonumber
\E_{d_1}(f(d_1,d_2)) := &
\langle \phi(y_1),\mu_y \rangle - \langle \phi(x_1),\mu_y \rangle - \langle \mu_x,\phi(y_1) \rangle
\\&- (
\langle \phi(z_1),\mu_z \rangle - \langle \phi(x_1),\mu_z \rangle - \langle \mu_x,\phi(z_1) \rangle
)
\\
\E_{d_1,d_2}(f(d_1,d_2)) := &
\squaredMMDpop{x}{y} - \squaredMMDpop{x}{z}
\end{align}

We are now ready to derive the dominant leading term,$\zeta_1$, in the variance expression \eqref{eq:derivation_of_the_variance_ofMMD}.

\paragraph{Term $\zeta_1$}

\begin{align}
\zeta_1 :&= \operatorname{Var}(\E_{d_1}(f(d_1,d_2))) \\\nonumber
&= \EXoneYoneZone [(\langle \phi(y_1),\mu_y \rangle - \langle \phi(x_1),\mu_y \rangle - \langle \mu_x,\phi(y_1) \rangle
- (
\langle \phi(z_1),\mu_z \rangle - \langle \phi(x_1),\mu_z \rangle - \langle \mu_x,\phi(z_1) \rangle)^2]\\
& \qquad \qquad 
- (\squaredMMDpop{x}{y} - \squaredMMDpop{x}{z})^2
\end{align}
We note many terms in expansion of the squares above cancel out due to independence. For example  $\E_{y_1,z_1}[\langle \phi(y_1),\mu_y \rangle \langle \phi(z_1),\mu_z \rangle]-\E_{y_1}[\langle \phi(y_1),\mu_y \rangle]\E_{z_1}[\langle \phi(z_1),\mu_z \rangle]=0$. 

We can thus simplify to the following expression for $\zeta_1$
\begin{align}
\zeta_1=&
\EYone [\langle \phi(y_1),\mu_y \rangle^2] - \EYone [\langle \phi(y_1),\mu_y \rangle]^2
\\& \nonumber
+\EXone  [\langle \phi(x_1),\mu_y \rangle^2] - \EXone  [\langle \phi(x_1),\mu_y \rangle]^2\\ & \nonumber
+\EYone [\langle \mu_x,\phi(y_1) \rangle^2] - \EYone [\langle \mu_x,\phi(y_1) \rangle]^2\\ & \nonumber
+\EZone [\langle \phi(z_1),\mu_z \rangle^2] - \EZone [\langle \phi(z_1),\mu_z  \rangle]^2\\ & \nonumber
+\EXone [\langle \phi(x_1),\mu_z \rangle^2] - \EXone [\langle \phi(x_1),\mu_z \rangle]^2\\ & \nonumber
+\EZone [\langle \mu_x,\phi(z_1) \rangle^2] - \EZone [\langle \mu_x,\phi(z_1) \rangle]^2\\ & \nonumber
-2 (\EYone [\langle \phi(y_1),\mu_y \rangle \langle \mu_x,\phi(y_1) \rangle] - \EYone [\langle \phi(y_1),\mu_y  \rangle] \EYone [\langle \mu_x,\phi(y_1) \rangle])
\\& \nonumber
-2 ( \EXone [\langle \phi(x_1), \mu_y \rangle \langle \phi(x_1), \mu_z \rangle ] - \EXone [\langle \phi(x_1), \mu_y \rangle] \EXone [ \langle \phi(x_1), \mu_z \rangle ])
\\ & \nonumber
-2 ( \EZone [\langle \phi(z_1), \mu_z \rangle \langle \mu_x, \phi(z_1) \rangle ] - \EZone [\langle \phi(z_1), \mu_z \rangle] \EZone [ \langle \mu_x, \phi(z_1) \rangle ])
\\ & \nonumber
\label{eq:variance_difference_of2MMD}
\end{align}

We can empirically approximate these terms as follows:

\begin{align}
\zeta_1 & \approx 
\frac{1}{n(n-1)^2} e^T \tilde{K}_{yy}\tilde{K}_{yy} e - \left(\frac{1}{n(n-1)} e^T \tilde{K}_{yy}e \right)^2
\\ & \nonumber
\qquad +  \frac{1}{n^2m} e^T K_{xy}^TK_{xy} e - \left( \frac{1}{nm} e^T K_{xy}e \right)^2
\\ & \nonumber
\qquad +  \frac{1}{nm^2} e^T K_{xy}K_{xy}^T e - \left( \frac{1}{nm} e^T K_{xy}e \right)^2
\\ & \nonumber
\qquad +  \frac{1}{r(r-1)^2} e^T \tilde{K}_{zz}\tilde{K}_{zz} e - \left( \frac{1}{r(r-1)} e^T \tilde{K}_{zz}e  \right)^2
\\ & \nonumber
\qquad +  \frac{1}{rm^2} e^T K_{xz}K_{xz}^T e - \left(\frac{1}{rm} e^T K_{xz}e \right)^2
\\ & \nonumber
\qquad +  \frac{1}{r^2m} e^T K_{xz}^TK_{xz} e - \left(\frac{1}{rm} e^T K_{xz}e \right)^2
\\ & \nonumber
\qquad - 2  \left( \frac{1}{n(n-1)m} e^T \tilde{K}_{yy} K_{yx} e
- \frac{1}{n(n-1)} e^T \tilde{K}_{yy}e \times \frac{1}{nm} e^T K_{xy}e\right)\\ & \nonumber
\qquad - 2  \left( \frac{1}{nmr} e^T \tilde{K}_{xy}^T K_{xz} e
- \frac{1}{nm} e^T K_{xy}e \times \frac{1}{rm} e^T K_{xz}e\right)\\ & \nonumber
\qquad - 2  \left( \frac{1}{r(r-1)m} e^T \tilde{K}_{zz} K_{xz}^T e
- \frac{1}{n(n-1)} e^T \tilde{K}_{yy}e  \times \frac{1}{nm} e^T K_{xy}e \right)
\end{align}

\subsection{Equality of Derivations}

In this section, we prove that Equation~\eqref{eq:denominator_of_the_TestStatistics} is equal to the variance of the difference of 2 $\squaredMMDpop{x}{y}$ and $\squaredMMDpop{x}{z}$.

\begin{align}
\sigma_{XY}^2 + \sigma_{XZ} - 2\sigma_{XYXZ} &= \EYone \left[ \langle \phi(y_1),\mu_y \rangle^2 \right] - \EYone \left[ \langle \phi(y_1),\mu_y \rangle \right]^2 \\
& \qquad+ \EZone \left[ \langle \phi(z_1),\mu_y \rangle^2 \right] -  \EZone \left[ \langle \phi(z_1),\mu_y \rangle \right]^2 \nonumber \\
& \qquad - 2 \left( \EYone \left[ \langle \phi(y_1),\mu_y \rangle \langle \phi(y_1),\mu_x \rangle \right] -  \EYone \left[ \langle \phi(y_1),\mu_y \rangle \right] \EYone \left[ \langle \phi(y_1),\mu_x \rangle \right] \right) \nonumber \\
& \qquad -2 \left( \EZone \left[ \langle \phi(z_1),\mu_z \rangle \langle \phi(z_1),\mu_x \rangle \right] - \EZone \left[ \langle \phi(z_1),\mu_z \rangle \right] \EZone \left[ \langle \phi(z_1),\mu_x \rangle \right] \right)  \nonumber \\
&  \qquad + \EXone \left[ \langle \phi(x_1),\mu_y \rangle^2 \right] - \EXone \left[ \langle \phi(x_1),\mu_y \rangle \right]^2 \nonumber\\
& \qquad + \EYone \left[ \langle \phi(y_1),\mu_z \rangle^2 \right] - \EYone \left[ \langle \phi(y_1),\mu_z \rangle\right]^2 \nonumber \\
&  \qquad + \EYone \left[ \langle \phi(y_1),\mu_x \rangle^2 \right] - \EYone \left[ \langle \phi(y_1),\mu_x \rangle \right]^2 \nonumber\\
& \qquad + \EZone \left[ \langle \phi(z_1),\mu_x \rangle^2 \right] - \EZone \left[ \langle \phi(z_1),\mu_x \rangle \right]^2 \nonumber \\
&  \qquad -2 \left( \EXone \left[ \langle \phi(x_1),\mu_y \rangle \right] \EXone \left[ \langle \phi(x_1),\mu_z \rangle \right] \right) \nonumber
\end{align}
We have shown that Equation~\eqref{eq:denominator_of_the_TestStatistics} is equal to Equation~\eqref{eq:variance_difference_of2MMD}.

%% file: calibrationOftheTest.tex
\section{Calibration of the test}\label{sec:appendixB}

\begin{wrapfigure}{r}{0.3\textwidth}
\centering
\begin{tabular}{cc}
\begin{sideways} $\qquad \qquad$ Frequency  \end{sideways} &  
\includegraphics[width=0.3\textwidth]{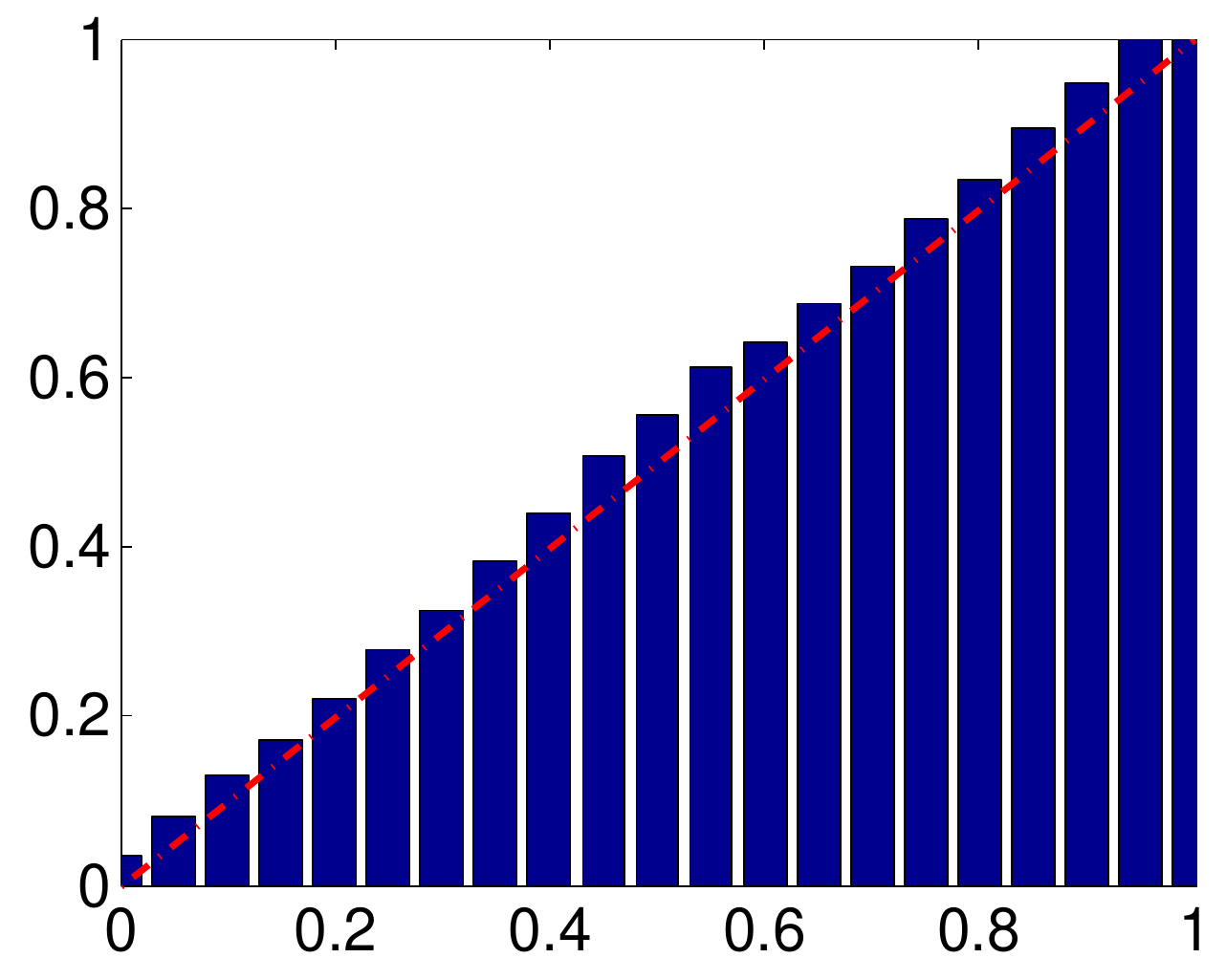} \\
& $p$-values \\
\end{tabular}
   \caption{Calibration of the relative similarity test}\label{fig:CalibrationDiagonalHist}
\end{wrapfigure} 

\begin{figure*}\centering

\begin{subfigure}[t]{0.5\textwidth}\centering
\includegraphics[width=0.6\textwidth]{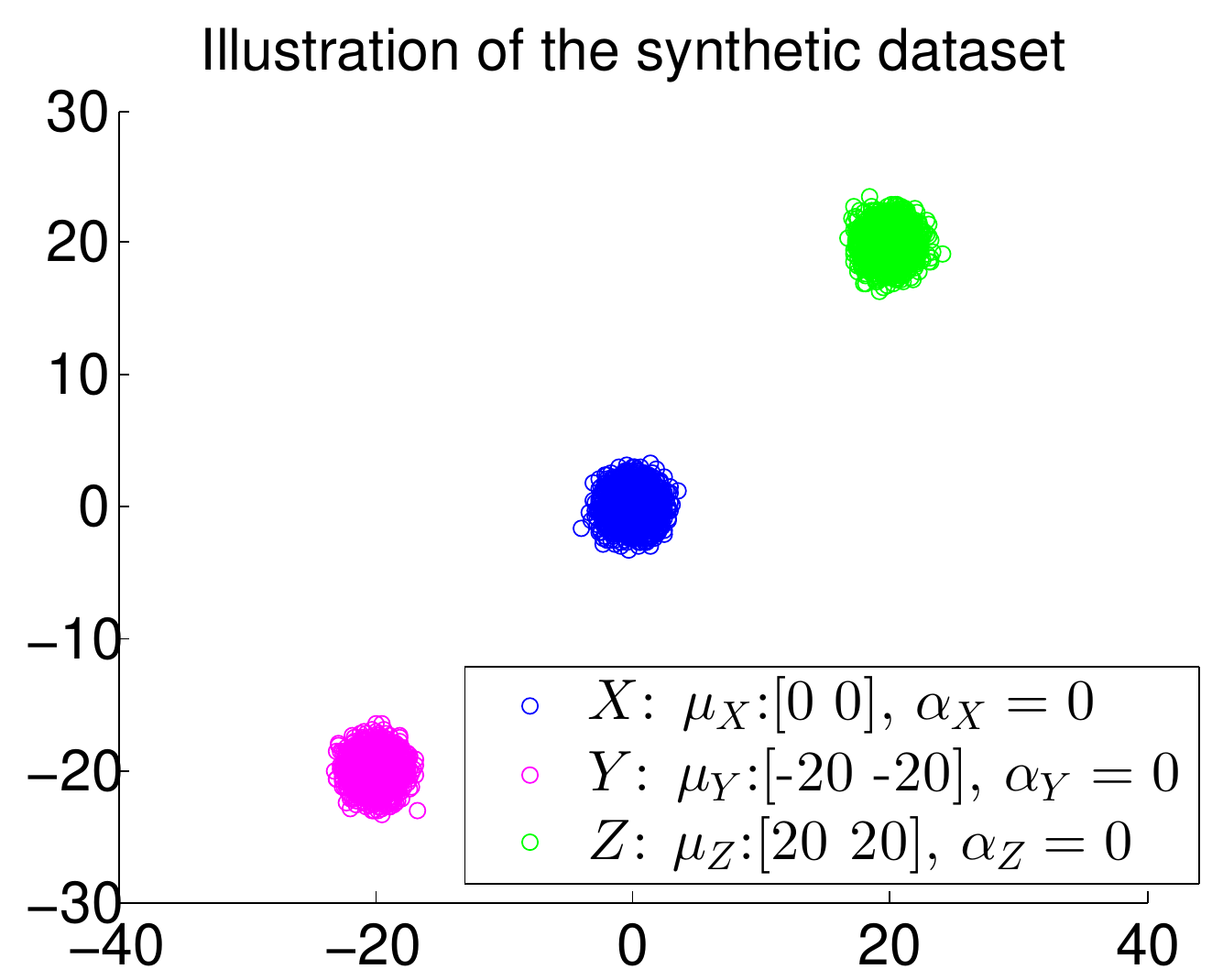}
\caption{Illustration of the synthetic data with different means for $X$, $Y$ and $Z$.}
\end{subfigure}%
\hfill
\begin{subfigure}[t]{0.5\textwidth}\centering
\includegraphics[width=0.6\textwidth]{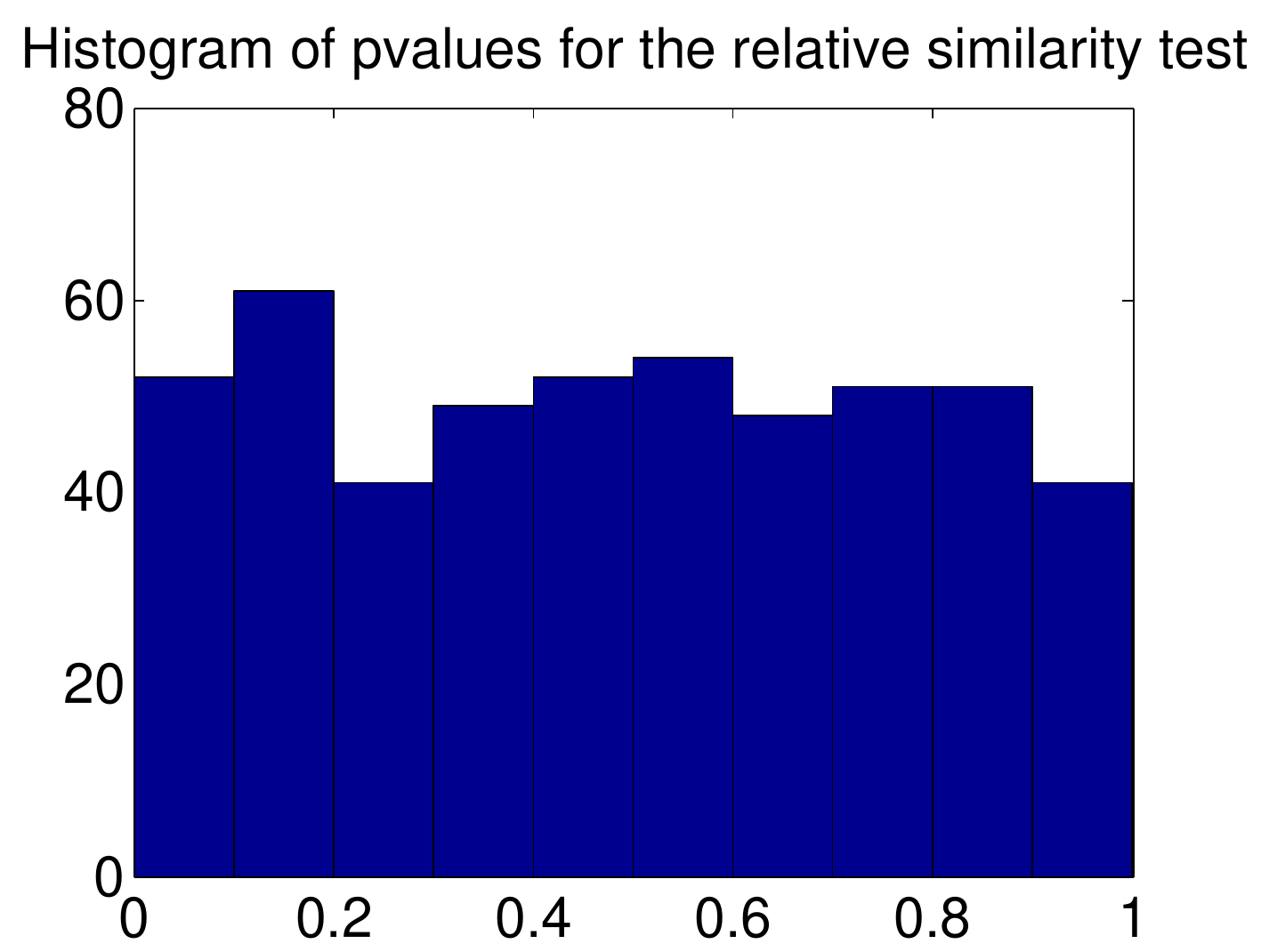}
\caption{Uniform histogram of $p$-values}
\end{subfigure}

\vskip \baselineskip

\begin{subfigure}[t]{0.5\textwidth}\centering
\includegraphics[width=0.6\textwidth]{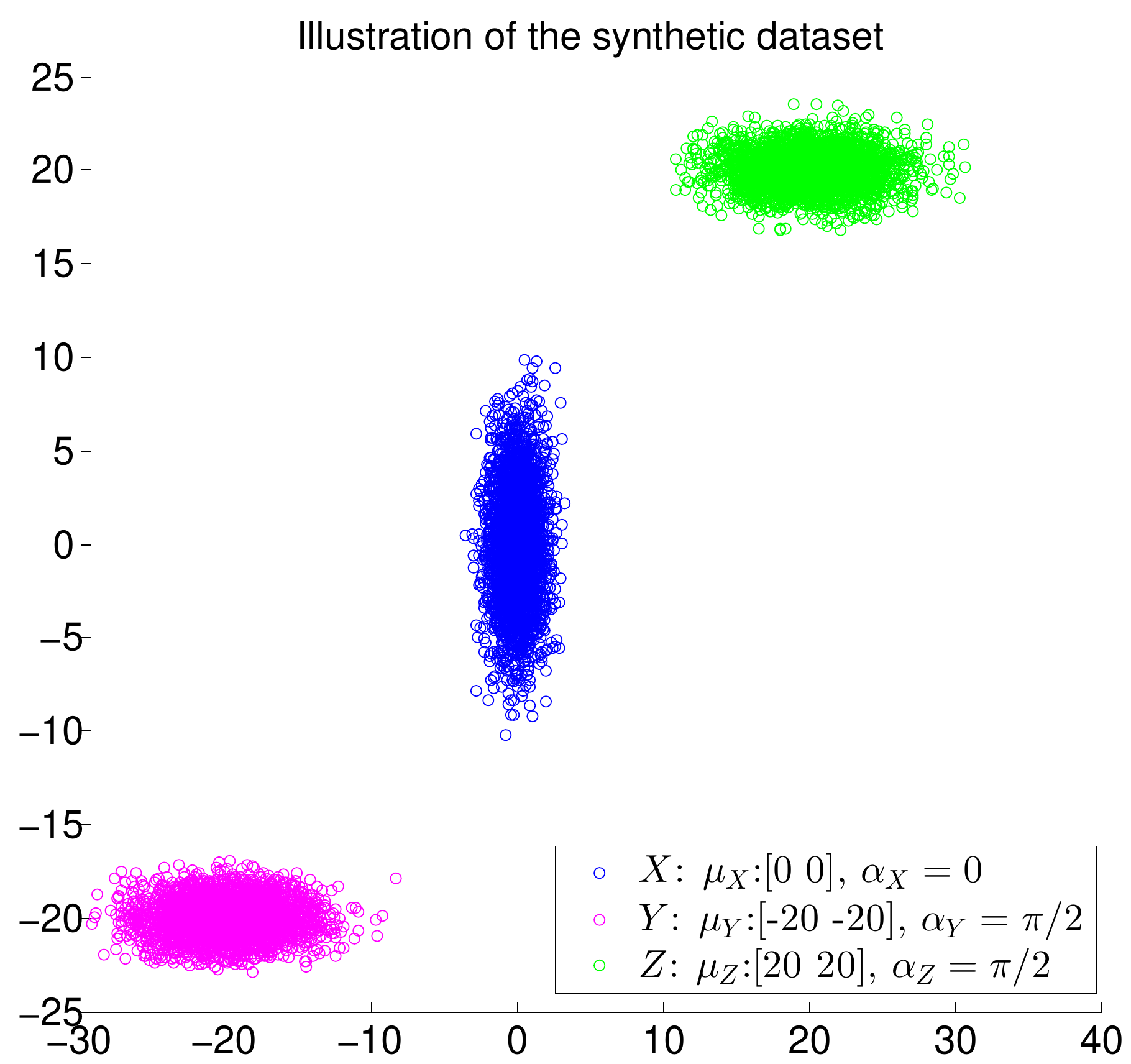}
\caption{Illustration of the synthetic data with different means and orientations for $X$, $Y$ and $Z$.}
\end{subfigure}%
\hfill
\begin{subfigure}[t]{0.5\textwidth}\centering
\includegraphics[width=0.6\textwidth]{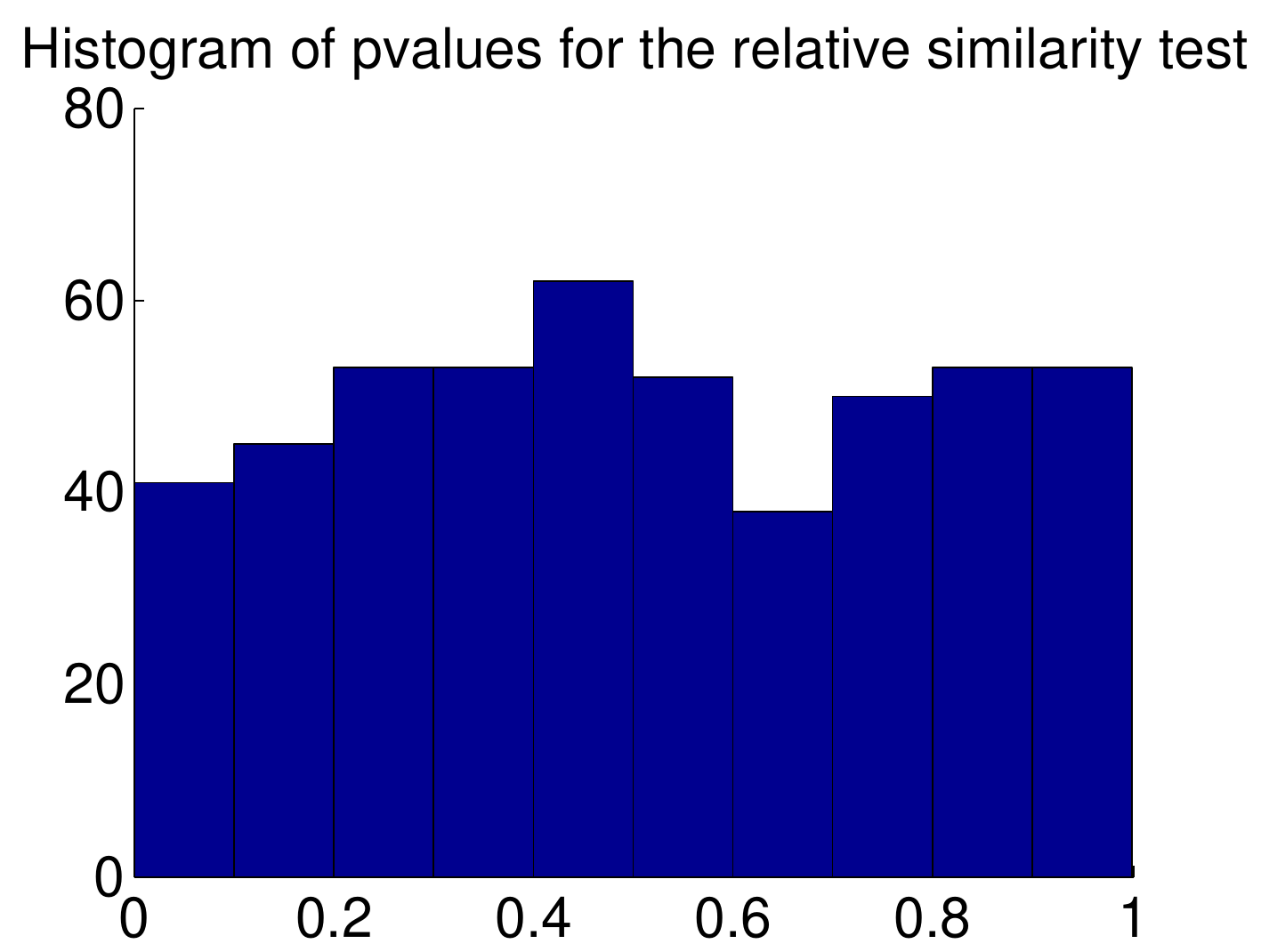}
\caption{Uniform histogram of $p$-values}
\end{subfigure}

\vskip \baselineskip

\begin{subfigure}[t]{0.5\textwidth}\centering
\includegraphics[width=0.6\textwidth]{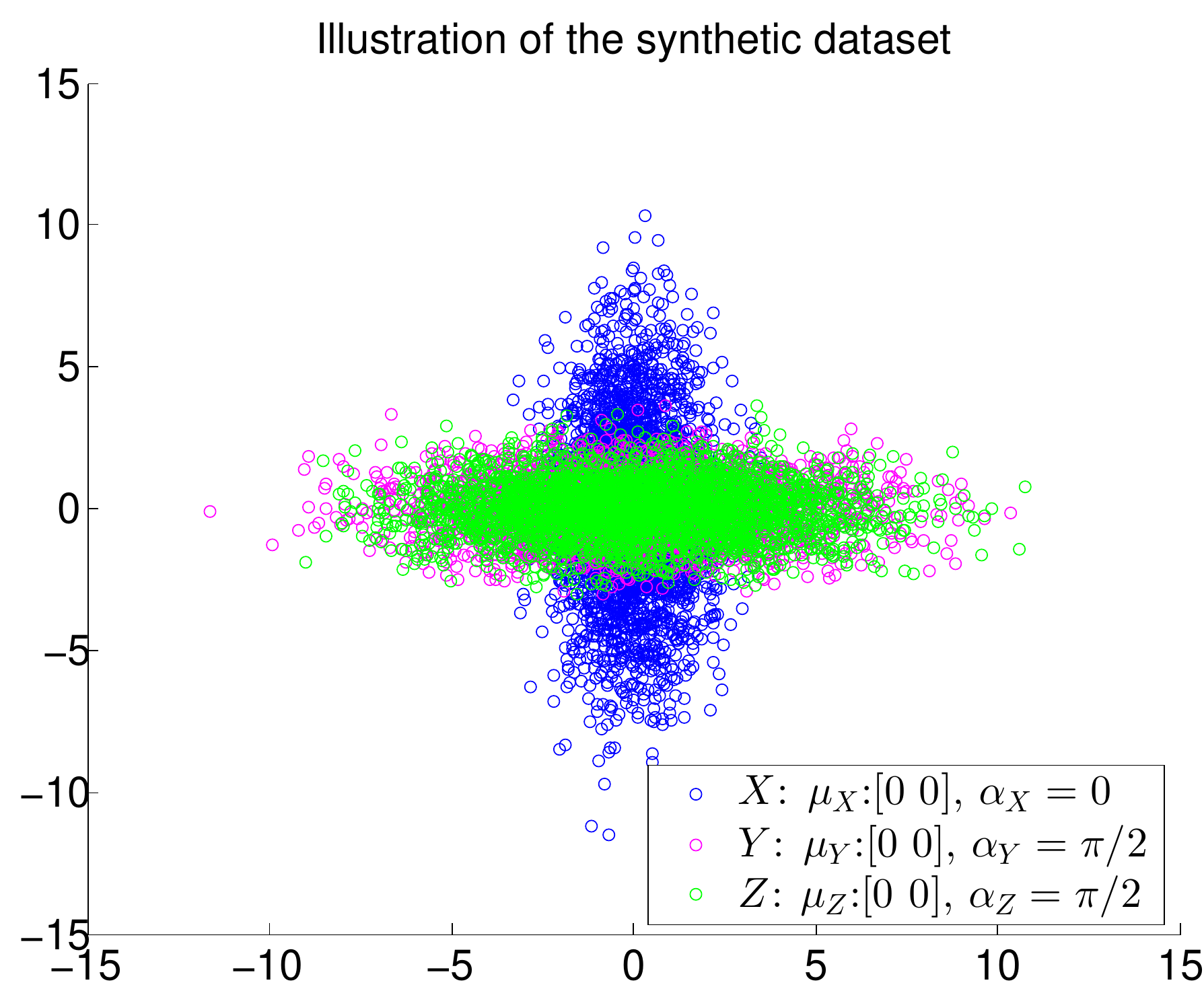}
\caption{Illustration of the synthetic data with different orientations for $X$, $Y$ and $Z$.}
\end{subfigure}%
\hfill
\begin{subfigure}[t]{0.5\textwidth}\centering
\includegraphics[width=0.6\textwidth]{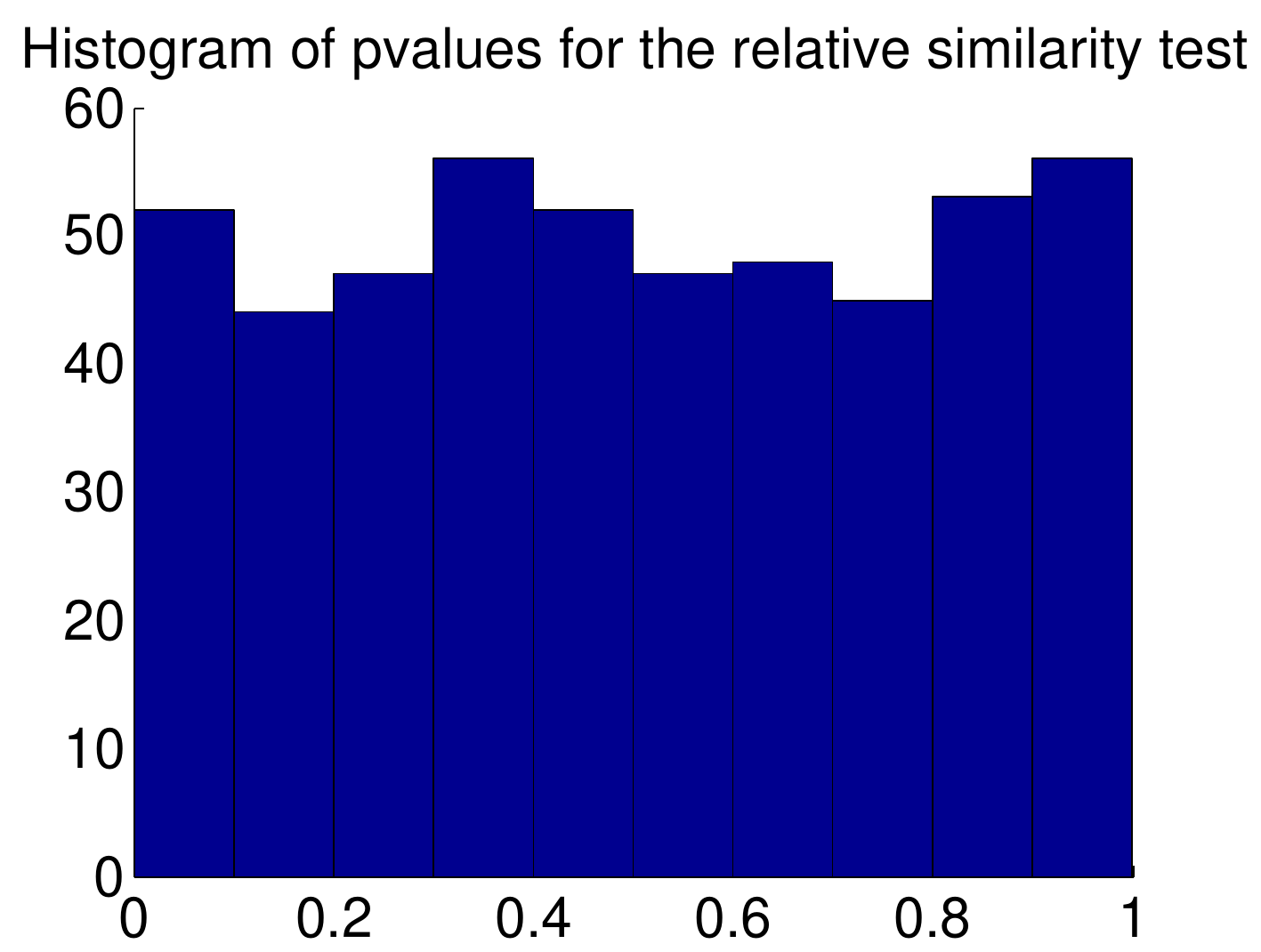}
\caption{Uniform histogram of $p$-values}
\end{subfigure}

\caption{Calibration of the relative similarity test}\label{fig:TestCalibrationUniformDist}
\end{figure*}

We show here that our derived test is well calibrated.  A calibrated test should output a uniform distribution of $p$-values when the two MMD distances are equal.  The empirical distributions of $p$-values for various sets of $P_x$, $P_y$ and $P_z$ are given in Figure~\ref{fig:TestCalibrationUniformDist}.  Similarly, for a given significance level $\alpha$, the false positive rate should be equal to $\alpha$.  The empirical false positive rates for varying $\alpha$ are shown in Figure~\ref{fig:CalibrationDiagonalHist} further demonstrating the proper calibration of the test.

